\documentclass[sigconf]{acmart}

\usepackage{graphicx}
\usepackage{geometry}
\usepackage{xcolor}
\usepackage{pifont}
\usepackage{enumitem}
\usepackage[utf8]{inputenc}

\usepackage{colortbl}   
\usepackage{xcolor}
\usepackage{subfig}

\usepackage{amsmath}
\usepackage{multirow}
\usepackage{soul}
\definecolor{Gray}{gray}{0.9}
\definecolor{LightGray}{gray}{0.95}

\definecolor{Gray}{gray}{0.9}
\definecolor{LightGray}{gray}{0.95}

\copyrightyear{2025}
\acmYear{2025}
\setcopyright{cc}
\setcctype{by-nc}
\acmConference[CHI '25]{CHI Conference on Human Factors in Computing Systems}{April 26-May 1, 2025}{Yokohama, Japan}
\acmBooktitle{CHI Conference on Human Factors in Computing Systems (CHI '25), April 26-May 1, 2025, Yokohama, Japan}\acmDOI{10.1145/3706598.3713306}
\acmISBN{979-8-4007-1394-1/2025/04}
\setlist[enumerate,itemize]{noitemsep, topsep=0pt}

\definecolor{RDcolor}{rgb}{0.5, 0.1, 0.8} 

\newcommand{\cmark}{\textcolor{green!50!black}{\ding{51}}} 
\newcommand{\xmark}{\textcolor{red}{\ding{55}}} 

\begin{document}



\def\name{3HANDS} 
\newcommand{\dataset}{\name}
\title[\name{}]{3HANDS Dataset:\\ Learning from Humans for Generating Naturalistic Handovers with Supernumerary Robotic Limbs}

\author{Artin Saberpour Abadian}
\affiliation{%
  \institution{Saarland University, \\Saarland Informatics Campus}
  \city{Saarbrücken}
  \country{Germany}}
\email{saberpour@cs.uni-saarland.de}

\author{Yi-Chi Liao}
\affiliation{%
  \institution{ETH Zürich}
  \city{Zürich}
  \country{Switzerland}}
\email{yichi.liao@inf.ethz.ch}

\author{Ata Otaran}
\affiliation{%
  \institution{Saarland University, \\Saarland Informatics Campus}
  \city{Saarbrücken}
  \country{Germany}}
\email{otaran@cs.uni-saarland.de}

\author{Rishabh Dabral}
\affiliation{
  \institution{Max Planck Institute for Informatics}
  \city{Saarbr\"ucken}
  \country{Germany}}
\affiliation{
  \institution{Saarland University, \\Saarland Informatics Campus}
  \city{Saarbrücken}
  \country{Germany}}
\email{rdabral@mpi-inf.mpg.de}

\author{Marie Muehlhaus}
\affiliation{%
  \institution{Saarland University, \\Saarland Informatics Campus}
  \city{Saarbrücken}
  \country{Germany}}
\email{muehlhaus@cs.uni-saarland.de}

\author{Christian Theobalt}
\affiliation{
  \institution{Max Planck Institute for Informatics}
    \city{Saarbr\"ucken}
  \country{Germany}}
\affiliation{
  \institution{Saarland University, \\Saarland Informatics Campus}
  \city{Saarbrücken}
  \country{Germany}}
\email{theobalt@mpi-inf.mpg.de}

\author{Martin Schmitz}
\affiliation{%
  \institution{Saarland University, \\Saarland Informatics Campus}
  \city{Saarbrücken}
  \country{Germany}}
\email{mschmitz@cs.uni-saarland.de}

\author{Jürgen Steimle}
\affiliation{%
  \institution{Saarland University, \\Saarland Informatics Campus}
  \city{Saarbrücken}
  \country{Germany}}
\email{steimle@cs.uni-saarland.de}



\renewcommand{\shortauthors}{Saberpour et al.}
\begin{abstract}
Supernumerary robotic limbs are robotic structures integrated closely with the user's body, which augment human physical capabilities and necessitate seamless, naturalistic human-machine interaction.
For effective assistance in physical tasks, enabling SRLs to hand over objects to humans is crucial.   
Yet, designing heuristic-based policies for robots is time-consuming, difficult to generalize across tasks, and results in less human-like motion. When trained with proper datasets, generative models are powerful alternatives for creating naturalistic handover motions.
We introduce 3HANDS, a novel dataset of object handover interactions between a participant performing a daily activity and another participant enacting a hip-mounted SRL in a naturalistic manner.
3HANDS captures the unique characteristics of SRL interactions: operating in intimate personal space with asymmetric object origins, implicit motion synchronization, and the user’s engagement in a primary task during the handover. 
To demonstrate the effectiveness of our dataset, we present three models: one that generates naturalistic handover trajectories, another that determines the appropriate handover endpoints, and a third that predicts the moment to initiate a handover. 
In a user study (N=10), we compare the handover interaction performed with our method compared to a baseline. The findings show that our method was perceived as significantly more natural, less physically demanding, and more comfortable.
\end{abstract}



\maketitle

\begin{figure*}[!t]
\centering
  \includegraphics[width=\textwidth]{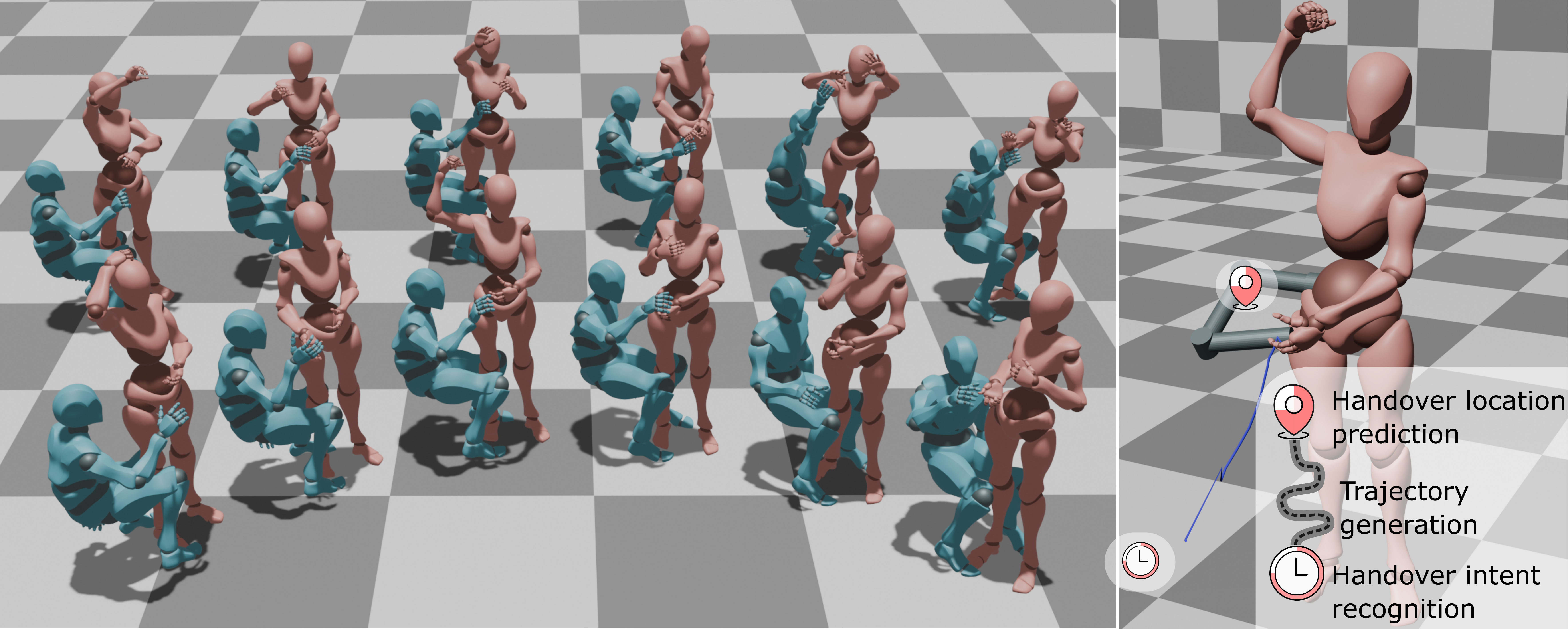}
  \caption{The 3HANDS dataset comprises an extensive collection of human motion data of asymmetric object handovers between users and a human-enacted third arm, which assists an ongoing activity by handing in or taking away objects at an intimate distance to the user.   
  It contains recordings of 946 interactions captured with 12 participant pairings while performing 12 daily activities. The dataset comprises rigged skeleton data of full body (69 joints) and hands (21 joints). We demonstrate the dataset's utility to train state-of-the-art machine learning models for three essential steps in the handover activity: generating naturalistic handover trajectories, predicting the location of the handover, and identifying the intent to initialize a handover.  }  
  \Description{this is the teaser figure showing people perform daily tasks}
  \label{fig:teaser}
\end{figure*}

\section{Introduction}

Supernumerary robotic limbs (SRLs) hold great promise in supporting humans in diverse activities by seamlessly integrating human bodies with assistive motion. 
Frequently investigated applications include physical activities where an additional hand is needed \cite{10187506ArashAjoudani-bidirectionalHandover}, physical assistance for the elderly \cite{parietti2015balance-augmentation}, augmenting humans with "superpowers" \cite{vatsal2018-srl-augment-superpower}, or assistance for strenuous physical tasks \cite{tong2021-SRL-augment, prattichizzo2021-srl-augment2, hussain2016-srl-augment3,9384151-srl-balance-above-head}. 
Prior works also found SRL's wide-range applications in the Human-Computer Interaction (HCI) field, such as holding support, offering additional control \cite{bodyIntegratedProgrammableJoints}, providing rich haptic feedback \cite{HapticPIVOT}, and complex human-robot collaboration \cite{vatsal2017wearing}.
In all these cases, handing over objects between human hands and the robotic limb is a frequent activity. 

Given that SRLs are human-machine interfaces that operate within personal and often even intimate distance to the user \cite{LEICHTMANN2020101386personal-distances, hall1966hidden}, their motion control demands are particularly stringent to ensure interactions that are predictable, safe, and effective. 
One might intuitively attempt to design handover motions for SRLs based on heuristics; however, this approach is a tedious control and programming task~\cite{Maeda2017PhaseEstimation}. It also can easily fail to account for the subtleties of natural human interaction, including naturalistic patterns of motion kinematics acceptable in intimate personal space, effective inter-hand motion coordination, and subtle cues that convey handover intention. 
Modern generative machine learning techniques offer a promising alternative, allowing us to develop data-driven models capable of generating natural and safe motions for user interaction~\cite{MoveInt,MILD}. Achieving this goal hinges on the availability of appropriate datasets for training these models. 

Several handover datasets currently exist. 
For instance, the H2O dataset~\cite{ye2021h2o} captures the hand postures at a short distance of both the giver and the receiver in front of each other, whereas the HOH dataset captures whole-body motions as two users sit face-to-face~\cite{wiederhold2024hoh}. Others captured bimanual handover motions~\cite{kshirsagar2023dataset}. 
While these datasets provide valuable insights into handover motions, they do not account for the differences in interactions and motion kinematics arising from operating in the user's personal space. Prior datasets studied face-to-face handovers with symmetric roles initiated based on clear temporal cues and without another primary activity. In contrast, the handover with an SRL is characterized by (1) asymmetric spatial configuration centered on the user's body, (2) asymmetric roles of SRL (assistant) and user (master), (3) ongoing primary activity of the user, potentially influencing timing and location of handover, and (4) implicit initiation of handover based on the user's implicit postural cues. 
These substantial differences demand a novel dataset that addresses these critical factors, which can then be used to train models for generating naturalistic SRL motions.

This paper contributes \dataset{}, a human-human object handover dataset specifically developed to help design the interactive behavior of hip-mounted supernumerary robotic limbs.  
It captures interacting pairs of humans (see \autoref{fig:teaser}). One person is performing 12 different daily activities, sampled from activities at the torso and above the shoulders, close or far from the body, and with a small to large range of motion, aiming to cover a broad spectrum of human motion dynamics. Examples range from shampooing hair and hammering to painting the wall and cleaning a window.
Concurrently, a second person acted as the SRL and was instructed to hand over and take back a spherical object to the first person in a natural manner, taking a position and posture representative of a hip-mounted SRL. 
%
We opted for a hip-mounted configuration, as it is a common SRL mounting location (e.g., \cite{parietti2015balance-augmentation,8962256,wrlkit}), more stable for mechanical movements~\cite{8962256}, minimizes interference with the user’s natural arm workspace, and enhances safety by distancing the SRL from sensitive areas (like the head and face).
We captured 946 interactions performed by 12 unique pairings of participants. The dataset was captured using a markerless motion-capture setup, with 41 synchronized 2K camera views. Markerless capture is considerably less invasive than the marker-based setups used in many prior studies and ensured that participants could move freely and naturally to perform the desired tasks without restraining their motions. 
The dataset contains detailed skeleton data of 69 joints with 107 Degrees of Freedom (DoF), including the detailed capture of hand articulation with 21 joints per hand. We recorded the participants' rigged 3D skeletons, hand poses, time-synchronized textual transcriptions of verbal utterances, information on whether a handover is occurring and the transcription of the verbal communication between participants with the time stamps for each interaction. 
%
%
%
Our dataset has four key characteristics that distinguish it from prior datasets and make it particularly proper for training SRLs: 
Contrasting with prior datasets that captured face-to-face handover with a clear handover temporal cue \cite{ye2021h2o,wiederhold2024hoh}, our dataset has the following features that made it particularly proper for training SRLs:
(1) Instead of face-to-face motions, handovers occur in an \textit{asymmetric spatial configuration} and \textit{in intimate distance to the user}, where the objects are asymmetrically delivered from the sides of the primary user. 
(2) Participants take on \textit{asymmetric roles}: primary user vs. robotic assistant. 
(3) The primary user performs an ongoing \textit{primary activity}.
(4) We opted against a specific cue after which both participants should initiate the motion immediately; instead, we precisely capture how participants \textit{implicitly coordinate} the start of a handover.  
This rich multi-modal dataset offers a valuable resource for the HCI community to investigate the complex interplay between human motion and verbal communication during handovers, ultimately informing the design of more intuitive and user-friendly SRL interfaces in particular and of human-robot interfaces in general. We share the dataset with the community\footnote{\href{https://hci.cs.uni-saarland.de/projects/3hands}{https://hci.cs.uni-saarland.de/projects/3hands/}}.


To further demonstrate practical applications of our dataset in training models for interaction with SRLs, we trained three distinct models using conditional variational autoencoder (CVAE) \cite{sohn2015CVAE} and neural network architectures. Each of them addresses one essential step in the handover activity. 
First, we developed a trajectory generation model capable of generating naturalistic handover motions for SRLs in response to the primary user's actions. 
Second, we contribute a model to anticipate the desired location where the handover will most likely occur for a given posture.
Finally, we show that our dataset facilitates the training of a model that accurately predicts when the SRL should initiate a handover solely based on implicit postural cues of the primary user. 
We detail on the data processing, models and experimental results. The performance metrics achieved with our dataset confirm its quality and show its potential to both, advance the field of SRLs and deepen the understanding of handover activities in close personal space. 
Furthermore, we conducted a user study examining the subjective perceived quality of the generated handover motions (for measures such as perceived naturalness, smoothness, and predictability) compared to a baseline method in a virtual reality environment. The results of the study indicated that our models trained with \dataset{} result in more natural and smooth motions that are less physically demanding and more comfortable. We hope our dataset and experimental results will provide a valuable resource for future studies and applications.

In summary, this paper contributes the following: 
\begin{itemize}
    \item We introduce the \dataset \space dataset, an extensive collection of motion patterns originating from two persons engaging in an object handover. It captures 946 asymmetric handover motions in scenarios where the user is performing a primary activity. It offers a rich set of motion data, comprising rigged 3D skeletons and hand poses, transcriptions of verbal utterances, and information on whether a handover is occurring.
    \item We illustrate the effectiveness of using the \dataset{} dataset to train models for handover interactions with supernumerary robotic limbs. These a) generate naturalistic handover motion trajectories, b) predict the location of a handover, and c) accurately predict when to initiate a handover.
    \item In a controlled user study in a virtual reality environment, we verify the naturalness of the handover interactions produced with a data-driven method trained on the \dataset \space dataset.
    \item We release the dataset to enable the community to create robust and reliable models of object handover with SRLs.
\end{itemize}

\section{Related Work} \label{chap:RelatedWork}

\subsection{Human-Human Handover}
In recent years, the study of human-human handover \cite{basili2009human-human-handover, parastegari2017modeling-human-human, strabala2013toward-seamless-human-robot-handovers} has gained attention due to its importance for improving human-robot interactions \cite{huber2008human-robot-interaction,gielniak2013generatingHumanLikeMotionForRobots} and collaborative systems \cite{someshwar2017collaborative-system-design,scimmi2019collaborative-robots}. 
Past works have investigated a wide range of factors influential to handover activities. These include the use of interpersonal space~\cite{hansen2017interpersonal}, timing~\cite{glasauer2010reaction-time}, 
handover context \cite{aleotti2014affordance}. They further addressed factors related to the handover objects, such as their physical properties  \cite{hansen2017object-mass, chan2020affordance, aleotti2014affordance, safetyAffordance}, 
associated gripping dynamics \cite{mason2005grip}, and
transfer control of the object~\cite{strabala2013toward-seamless-human-robot-handovers}. 
Other works have investigated giver and receivers' motions \cite{huang2015adaptive} to communicate intent before handover \cite{strabala2012intent-communication}, as well as social bonding and shared goals \cite{wolf2016joint-attention}. 
%
Building upon these rich insights, significant advancement has been made in data-driven control methods for human-robot handover~\cite{huang2015data-driven-adaptive, al2021data-driven-HRI, yamane2013data-driven-synthesizing,kshirsagar2019data-driven-control-HRI, khanna2022data-driven-grip}.
%
The data-driven approaches, which are trained on human-human handover data, have been shown to enable robots to better adapt to human behavior for smoother and more intuitive interactions~\cite{legibility_predictability, Sheikholeslami2020}. 

Several datasets have been developed to study human-human handovers, each varying in terms of setup, modalities, and object interactions. 
The HoH dataset \cite{wiederhold2024hoh} and the dataset by Khanna et al. \cite{khanna2023dataset} involve participants with a table between them, either seated or standing. HoH provides point clouds, while Khanna et al. include motion tracking along with handover forces. 
The H2O dataset \cite{ye2021h2o}, and the datasets by Kshirsagar et al. \cite{kshirsagar2023dataset} and Chan et al. \cite{chan2020dataset} involve participants standing at a comfortable distance, with variations in their sensor setups: H2O employs magnetic sensors and cameras to focus on hand dynamics, while the others utilize markered motion capture, RGB-D data, and multiple camera views. 
Carfi et al. \cite{carfi2019dataset} provide a more dynamic scenario, with participants freely moving toward each other in various handover contexts, incorporating multimodal data like motion capture, IMU, and videos. 
Lastly, Cini et al. \cite{cini2019dataset} focus on grasps used during hand-object interactions.

To the best of our knowledge, existing datasets for human-to-human handovers are limited in that they focus on symmetric constellations where the giver and receiver face each other, and the handover occurs centrally in their shared interpersonal space. Furthermore, these datasets lack the implementation of an ongoing activity that is performed before and after the handover. 
To address these, we propose \dataset{} that is focused on an asymmetrical giver-receiver relationship in close peripersonal space while the primary user is also engaged in an activity.
A comparison of the datasets is presented in the Table~\ref{table:handover-dataset}. 
\begin{table*}[t] 
\centering
\resizebox{\textwidth}{!}{ 
\begin{tabular}{l|c c c c c c | c }
\textbf{Dataset} & \textbf{Cini \cite{cini2019dataset}} & \textbf{Chan \cite{chan2020dataset}} & \textbf{Carfi \cite{carfi2019dataset}} & \textbf{Kshirsagar \cite{kshirsagar2023dataset}} & \textbf{H2O \cite{ye2021h2o}} & \textbf{HOH \cite{wiederhold2024hoh}} & \textbf{\dataset ~(ours)} \\
\hline\hline
\textbf{Human-human spatial zone} & - & Social & Social/public & Personal/social & - & Personal/social & Close intimate/intimate  \\
\rowcolor{LightGray}
\textbf{Activities} & \xmark & \xmark & \xmark & \xmark & \xmark & \xmark & 12 \\
\textbf{Interactions} & 1734 & 1200 & 288 & 240 & 1200 & 2720 & 946 \\
\rowcolor{LightGray}
\textbf{Unique participant pairings} & 17 & 10 & 18 & 24 & 40 & 40 & 12 \\

\textbf{Markerless} & \xmark & \xmark & \xmark & \xmark & \xmark & \cmark & \cmark \\
\rowcolor{LightGray}
\textbf{Cameras} & 1 & 8 & 1 & 2 & 5 & 8 & 41 \\

\textbf{Full body 3D skeleton} & \xmark & \xmark & 9 joints & 13 joints & \xmark & \xmark & 69 joints \\
\rowcolor{LightGray}
\textbf{Hands 3D skeleton} & \xmark & \xmark & \xmark & \xmark & \xmark & \xmark & 21 joints per hand \\
\textbf{Objects} & 17 & 20 & 7 & 5 & 30 & 136 & 3 \\

\rowcolor{LightGray}
\textbf{Experimental Validation} & \xmark & \xmark & \xmark & \xmark & \cmark & \cmark & \cmark \\
\multirow{ 2}{*}{\textbf{Setting}} & \multirow{ 2}{*}{-} & Standing,  & Standing, & Standing,  & Standing,  & Seated, & Standing-seated, \\
 &  & Freely moving & Freely moving  & Face-to-face &  Face-to-face & Face-to-face  & Asymmetric \\
\rowcolor{LightGray}
\textbf{Suitable for SRLs} & \xmark & \xmark & \xmark & \xmark & \xmark & \xmark & \cmark \\

\end{tabular}
}
\caption{Comparison of \dataset{} with  prior human-human handover datasets. Human-human spatial zones are inferred based on Lambert's definition of spatial zones \cite{lambert2004body-spatial-zones}.}
\label{table:handover-dataset}
\end{table*}
\subsection{Human-Robot Handover Control}

Human-robot handover tasks combine anticipation of human intent with path-planning algorithms to generate feasible and natural handover trajectories. Generated trajectories optimize safety, reachability, and timing, to ensure smooth and collision-free handovers. 
Models on natural reaching movement, such as the minimal jerk model \cite{Hogan1982} have been used for anticipating handover timing and location \cite{Li2015,Landi2019}. Elliptic trajectory modeling \cite{Sheikholeslami2020} was proposed for making early and fast predictions on the movement of the collaborator and was shown to perform better than the minimal jerk model \cite{Chan2020}. 
While classical modeling approaches provide fast computation times and hard constraints to ensure safety, they are prone to model inaccuracies and need more tuning effort from an experienced designer in custom scenarios.

In addition, classical modeling approaches are not well-suited to account for the subtleties and multimodality of naturalistic human interactions. 
Prior work on human-robot proxemics has highlighted the relevance of personal spatial zones for human-robot interaction \cite{1513803-proxemics1, takayama2009influences-proxemics3} and balanced physical distancing \cite{argyle1965equilibrium}. It has been shown that robots must follow societal norms of physical distancing to offer smooth and comfortable, rather than disruptive and threatening interactions \cite{mumm-proxemics2}.
Spatial invasion, due to inappropriate distances between the robot and the human, can result in discomfort and avoidance \cite{leichtmann2020mobile-robot-interpersonal}.


%
%



Data-driven or hybrid approaches are more capable of capturing subtle dynamics. The number of applications that rely on such methods is increasing as the availability of handover datasets improves. These approaches can be used to tune specific model parameters \cite{Medina2016}, make real-time predictions \cite{Yang2022ModelPC, Maeda2017PhaseEstimation}, or control the entire process using generative models \cite{MoveInt}. In this paper, we show that our \dataset{} dataset provides high-quality and detailed human pose data to enable the training of generative handover control architectures enabling naturalistic and fluid handovers without basic heuristic constraints. Furthermore, our work contributes to future analyses of personal space in human-robot interaction, an important area that is still in its infancy \cite{LEICHTMANN2020101386personal-distances}. 


Handover control for supernumerary robotic limbs has received significantly less attention than for stationary robots. Existing solutions either rely on human input by utilizing human redundant degrees of freedom \cite{metaArms} or use heuristic methods \cite{10187506ArashAjoudani-bidirectionalHandover}. The lack of more data-driven approaches for wearable interfaces can be attributed to a lack of available datasets for training, imitating natural handover scenarios with an agent that resides in the user's personal space. We address this problem by contributing a comprehensive dataset focused on movement configurations that are specific to supernumerary robotic limbs.

\subsection{Supernumerary Robotic Limbs}

In recent years, wearable robotics have emerged as an expanding topic of study. These include supernumerary robotic limbs that augment users by providing additional extra limb-like robotic structures~\cite{superNumeraryRoboticLimbsOverview}, prosthetics that replace missing body parts~\cite{prostheticLimbsOverview}; and exoskeletons that help to improve the physical performance of the user's existing limbs~\cite{exoskeletonsReview}.

Supernumerary robotic limbs have been extensively researched, primarily in the robotics literature but also increasingly in HCI. Numerous structural configurations have been suggested by researchers for Supernumerary robotic limbs. For instance, prior work~\cite{locationForeArm} proposed a forearm-mounted supernumerary robot, dexterous torso-mounted robotic arms~\cite{sasaki2017metalimbs}, a shoulder-mounted extra arm for above-the-head work~\cite{aboveTheHeadPetriNet}, or additional finger-like structures~\cite{bodyIntegratedProgrammableJoints}. A pliable snake-shaped wearable robot featuring 25 degrees of freedom has been developed for highly adaptable application to the body in various geometric arrangements~\cite{orochi}. Various end-effectors are also suggested for SRLs ~\cite{teroos}.
Beyond physical assistance, SRLs are promising for virtual reality~\cite{EmbodimentOfWRLInVR} and haptics, where wrist-worn~\cite{HapticPIVOT} or waist-worn~\cite{HapticSnakes} robots can offer rich haptic feedback on multiple body locations. 
Another line of work investigates the important challenge of how to adapt an SRL to individual bodies of users and to individual body locations~\cite{Tomura}. Key directions include creating customized SRLs by assembling modular hardware building blocks~\cite{MorphologyExtensionKit} or using motion capture, digital design, and optimization algorithms to digitally customize a device design for computational manufacturing~\cite{wrlkit}. 
A central question involves how to control the motion of SRLs. 
To manage the motion trajectories of the SRLs, researchers have investigated the interactions between the user and the device~\cite{iNeedaThirdArm}. This is a hard challenge because, when operating an SRL, the user's body is frequently occupied with a primary manual activity, restricting conventional touch or gesture-based interaction.  
One line of inquiry centers on robot planning, which employs activity recognition to autonomously steer the robot toward a goal that negates the need for direct human interaction~\cite{DemonstrationBasedControlOfSupernumeraryRoboticLimbs}. Another line of inquiry uses remapping of body motion, where degrees of freedom in body movement that are not required for a specific task are remapped to control the SRL. For instance, mapping a user's foot movements to robotic arms can be a promising technique for intuitive and flexible real-time control~\cite{metaArms}. 
Other approaches proposed using the back of the hand~\cite{assistiveSupernumeraryGraspingWithBackOfHand}, the pinky finger~\cite{mechanicalDesignOfSupernumerary}, or capturing muscle movements with EMG~\cite{mappingHumanMuscleForceToSupernumeraryRoboticsDeviceForOverheadTaskAssistance} for controlling an SRL. 
Our work contributes to interactions with SRLs by proposing to learn subtle and nuanced motion dynamics from pairs of interacting humans.

\section{Dataset} \label{chap:Dataset}

In this section, we detail on the \name{} dataset, an extensive collection of motion patterns originating from two persons engaging in an object handover. It comprises detailed motion data of more than 946 interactions where the primary person is performing 12 daily activities while the second person is enacting a hip-mounted third arm that hands over and takes back objects to assist the primary person during the daily activity. 
The decision to use a hip-mounted SRL in \name{} is based on its popularity in related work and its ability to minimize interference with the user's natural active space compared to other common SRL mounting locations. Additionally, it enhances safety by keeping the SRL away from sensitive areas such as the head and face.
By recording interactions that were intuitively performed by two interacting humans, the dataset captures the specific and mostly implicit requirements of operating in intimate personal space as well as the interpersonal dynamics of object handover during a primary activity.

 
%
%
%

\subsection{Apparatus and Captured Motion Data}

We recorded the participants using the markerless optical motion capture system \emph{Captury}\footnote{\url{https://captury.com}}, which is based on the skeleton tracking approach of Stoll et al.~\cite{stoll} with additional hand tracking and a comparable average range of error of 8.79~mm compared to the marker-based Vicon system (cf. \cite{harsted2019performance}).
The allocentric setup uses 41 time-synchronized RGB cameras mounted at the walls and ceiling, each recording at a resolution of $2056$ × $1504$ pixels with $25$ Hz framerate. 
This multiview motion capture effectively minimizes data loss caused by occlusions, as occluded joints are likely visible in other views.
As extreme occlusions could cause challenges similar to marker-based systems, our manual verification confirmed the motion quality without any instances of mistracked joints.
They capture the motion of multiple persons simultaneously, in an area of $7 \times 6$~m.
%
In a \textit{markerless} motion-capture setup, the participants are not required to wear body suits or stick optical markers on their bodies.
%
%
This \textit{non-invasive} capture setup allows the participants to freely and naturally perform the desired tasks with no restrictions on the kind of motions they can exhibit.
Additionally, the system also provides tracking of finger joints, thereby allowing us to capture fine-grained handover.
Not having markers facilitates better capture of such finger articulations as it is typically difficult to attach and label markers on the fingers. 
\par
The setup provides rigged skeleton data of both interacting participants, including their hand poses ($21$ joints for each hand). 
As human joints have limits on the articulation angle and not all joints rotate along all three axes, the mocap system defines the skeleton as a kinematic tree of $107$ \textit{Degrees-of-Freedom} (DoF).
Each DoF represents the axis-aligned rotation of a joint along a specific axis defined in the local coordinate system.
The DoFs are also assigned individual limits on the maximum and the minimum articulation based on statistical data.
These skeleton DoFs can be transformed into body joint rotations represented using Euler angles (or quaternion).  
Further, we perform a Forward Kinematics operation on the joint angles to recover the 3D positions of each body joint.
 In total, our skeleton definition comprises 69 joints and 107 Degrees of Freedom (DoF).



%

We also recorded the audio of the spoken instructions provided by the primary participant using an omnidirectional neckband microphone attached to the participant.
In order to synchronize the audio with the captured motion, we ask the participant to clap three times at the start and the end of the recording sequence.
The peaks at the audio channel and clap moments of the hand joints are then aligned to achieve synchronization.

\subsection{Activities}
\label{sec:Activities}
Since we aim to capture a broad spectrum of motion patterns, we let participants perform 12 manual activities, representative of everyday activities that frequently require object handovers. 
In order to select activities that represent a broad set of tasks, we  systematically select such activities that provide a large coverage along the  following parameters of
motion patterns: 1) \textbf{Height} at which the action occurs relative to the user's body; we include tasks at the level of the user's \textit{torso} and \textit{head}. 2) \textbf{Distance from the body}: we vary activities carried out \emph{on-body} (these require particularly careful motion to avoid uncomfortable or even hurtful encounters) and \emph{mid-air} activities carried out at a certain distance in front of the body. 3) \textbf{Motion range}: we distinguish between \emph{small} motion range, where hands stay largely at the same location for performing dexterous tasks (e.g., adjusting a small picture at the wall), \emph{medium} range of motion (e.g., hammering a nail), and \emph{large} range of motion (e.g., painting a large wall).

%

Based on the parameters defined above, we select 12 everyday manual activities that each cover a unique combination in this $2 \times 2 \times 3$ parameter space. \autoref{tab:activites} depicts the set of activities.
We provided the participants with authentic props for each task to enhance the realism of performing the activities. For instance, a hammer was provided for hammering, a washcloth for washing the torso, etc.
Additionally, we recorded a neutral pose where users were instructed to comfortably rest their arms while standing still. 


%

%

\begin{table*}
 \includegraphics[width=.9\textwidth]{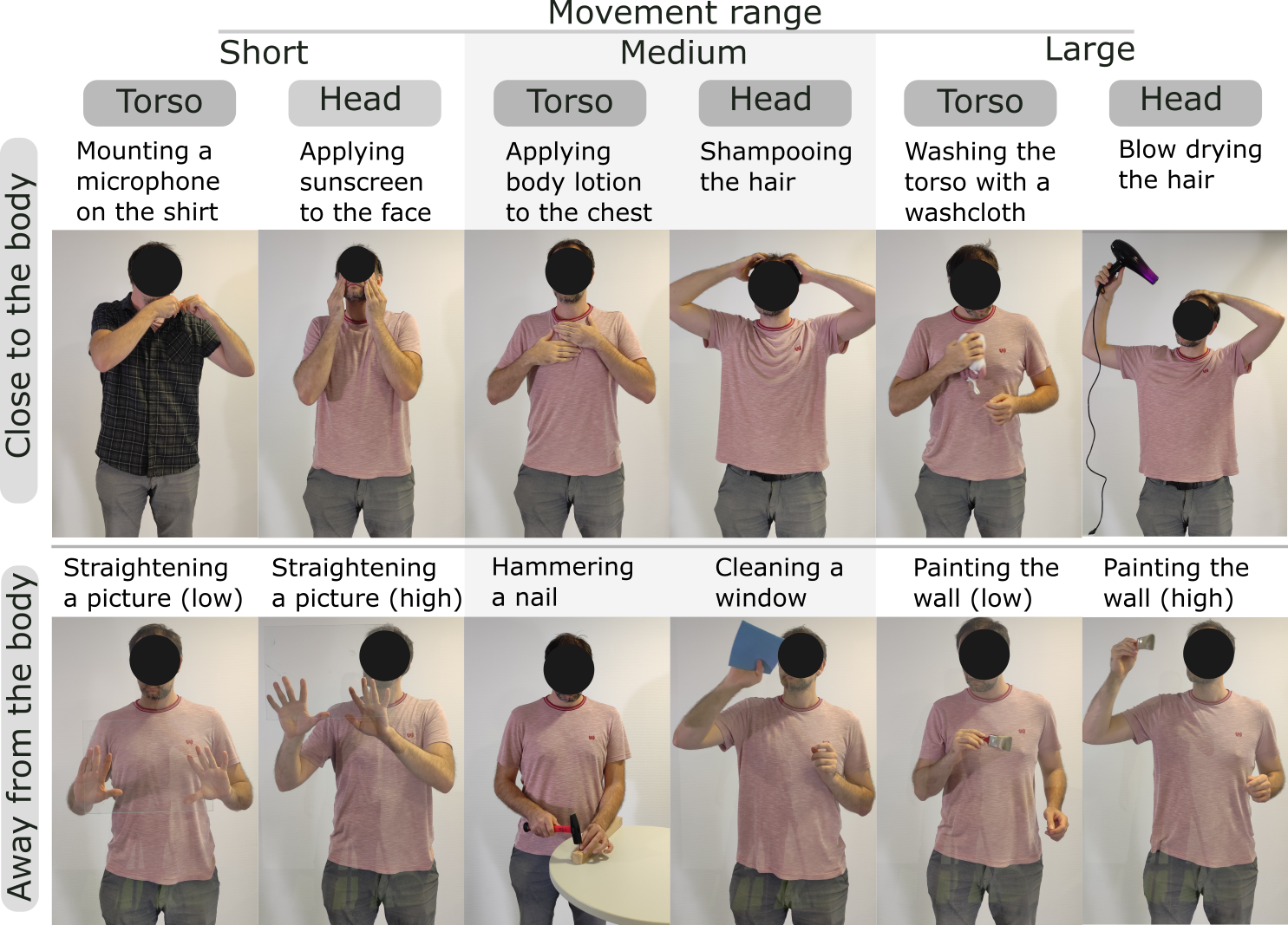}
	\caption{The set of activities captured in the dataset. Transparent acrylic sheets were used for the wall and the picture frame to avoid visual occlusions.}	
        \label{tab:activites}
\end{table*}

\subsection{Task and Procedure}

\subsubsection{Roles and Spatial Setup}
\begin{figure}[t!]
    \centering
    \includegraphics[width=1.0\linewidth]{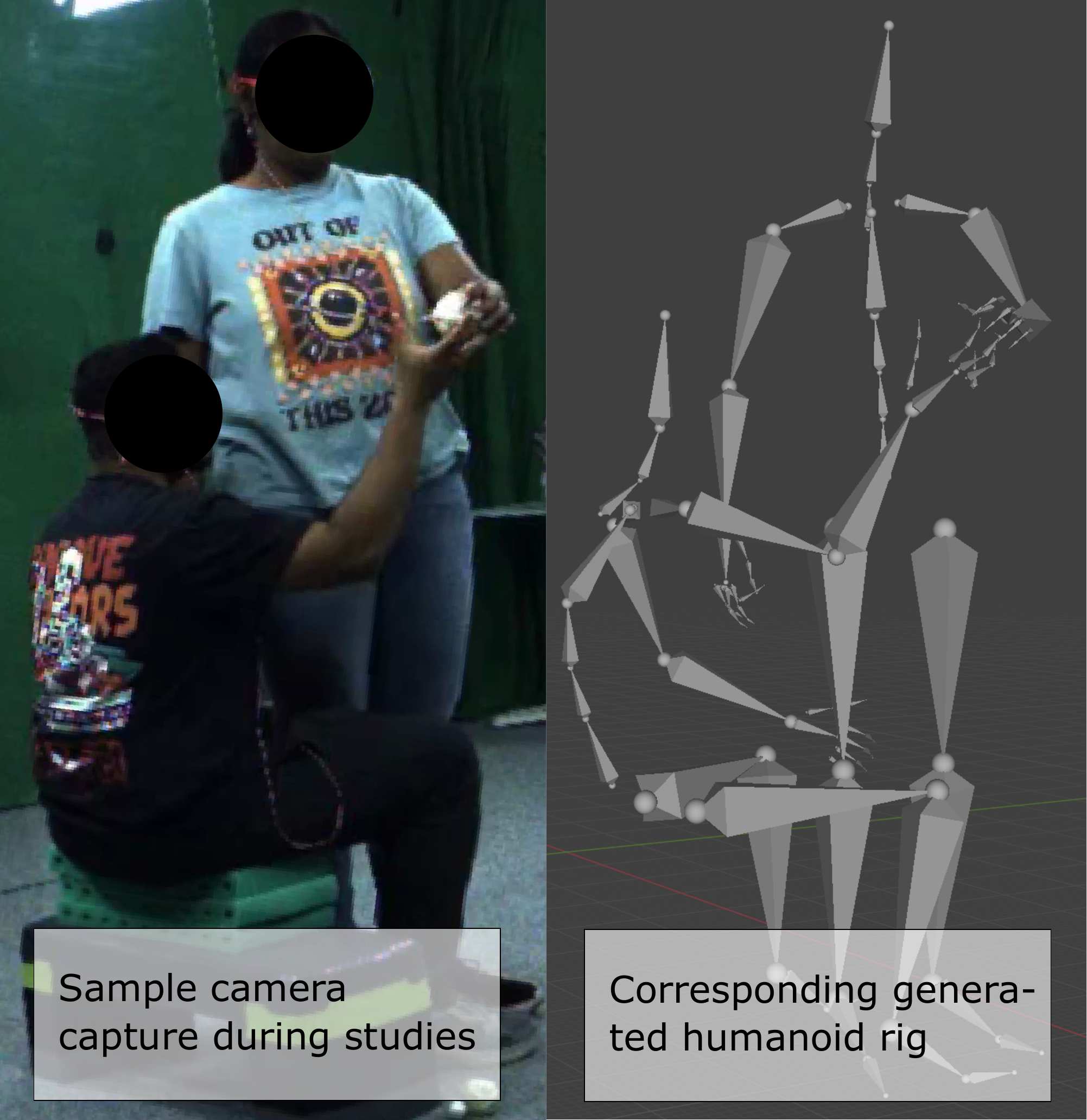}
    \caption{Setup of the asymmetric handover task. Left: the primary participant was standing and performing the primary activity, while the second participant enacted a robotic arm for handing over an object. Right: we generate rigged skeletons of both humans, including their articulated hands.}
    \label{fig:participantsConfig}
\end{figure}

One participant takes the role of the user (called the \emph{primary participant}), and the other one acts as the serving robotic arm (called the \emph{robot participant}). 
Pairs were instructed that the robot participant should aim at assisting the primary participant to the best level possible in handing over and taking objects, while the primary participant should focus on the primary activity and not care about the robot participant. 
Contrary to previously introduced handover datasets \cite{wiederhold2024hoh, ye2021h2o, carfi2019dataset, kshirsagar2023dataset, chan2020dataset, cini2019dataset}, our participant pairs do not face one another. Instead, we intentionally arranged the setup to resemble a hip-mounted SRL on the dominant hand's side. Therefore, the robot participant was asked to sit on the dominant hand side of the primary participant where the shoulder of the robot participant is at the hip level of the primary participant, facing the primary participant's hips at a slight distance (approx. 20cm), so as to not block the primary participant's elbow while performing the activity. The primary participants were standing and wore glasses that shielded their peripheral view on the lower right. This shielded the robot participant's face from their peripheral view, enabling them to focus on their activity and avoid communicating through eye contact.
We illustrate the arrangement of the participants in \autoref{fig:participantsConfig}.

%

The height of the stool for the robot participant is adjusted such that the robot participant's shoulder is aligned with the height of the primary participant's hip level. 
For the activities that required to be performed on a wall, we provided a wall-sized fixed acrylic panel in order to not block the camera views. 
\subsubsection{Handover Task}
The experimenter first communicated the general instructions by playing a voice recording. 
After a short trial run, the pairs then performed the following handover task for each of the 13 activities (12 + 1 neutral pose, activities were performed in randomized order):
After hearing a beep sound, the primary participant is performing the activity with the provided prop object, standing upright. The robot participant's right arm is in a resting position (hanging down), holding the handover object. Next, the primary participant initiates a handover at any preferred time. The primary participant is free in the modality and way they would want to make the robot participant aware of their intention for handover. 
Then, the robot participant starts handing the object over to the primary participant and then goes back to the resting position. The participants were instructed to perform this motion in a way they considered natural. Once the object is handed over, the primary participant briefly mimics using the object. At any preferred time, the primary participant then signals the robot participant to take away the object. The robot participant's right arm starts moving, takes the object, and returns to its resting position. 

Since we solely focus on the motion patterns and not on the specifics of the grasp, we assigned a fixed handover object per activity that is chosen among three spherical objects with diameters 2.5, 4, and 6cm, assigned with relevance to handover objects in the scenarios.

%

\subsubsection{Trials}

The participants repeated the handover three times for each activity. After this process, the pair reversed roles and fully repeated it again, resulting in 156 trials (12+1 activities x 2 hand-to/take-away x 3 repetitions x 2 reversed roles). In 16 instances, participants have performed 4 instead of 3 repetitions. Data from 7 trials had to be discarded due to the primary participant looking at the robot participant's face or the robot participant's right hand not waiting in the rest pose. In summary, the dataset contains 946 captured interactions.
For one pair of participants (including reversed roles), the whole capturing session took approximately 90 minutes.

\subsection{Participants}
We recruited 12 participants (6 male, 6 female, 0 non-binary), aged between 21 and 33 years old. 
All participants reported being right-handed. They received a monetary compensation. 
Participants conducted the data capture in pairs and reversed roles to double the number of unique pairings.  
Since our setup requires operating in the intimate peripersonal space around the body, we opted for recruiting only couples who are in a stable relationship. 
To ensure that the arm's length while sitting is sufficient to reach the location for object handover, we only included couples whose difference in height was not more than 20~cm. 

\subsection{Data Processing}
The output of the preprocessed data is the motion data of the interacting participants. It includes the motion files (in BVH and FBX format) together with time-synchronized raw video of all 41 cameras and the audio recording. 

We then manually annotated the Ground Truth (GT) by marking the frames that belong to a handover activity. We define a handover activity to begin when the robot user starts moving; it ends when the robot user’s hand has returned to its resting position after the object has been handed over. Each handover frame is annotated with the correct label of the specific handover task (give to or take away). In every frame of a handover, we furthermore labeled whether the object was in the primary participant's hand or in the robot participant's hand. The timestamp for every valid frame in its relevant handover segment is also added to the data (time data). This can be used later to provide temporal information about the current status of the interaction with the model. 
%
%

For each frame of a handover, after marking the start and end times of the GT segments, we store the 3D rotations (in parent-relative coordinates) and calculate and restore the 3D positions (in global coordinates) of each joint of both participants for all the valid frames within the sequence. 
We also include the stored rotation and position joint values in the dataset (CSV format). 
The 3D joints' positions' calculations are based on the root joint’s location and rotation (hip joint of each participant), following the skeletal hierarchy and using the local rotations and segment lengths specified in the motion files. 
We also provide the transcription of the associated audio files including verbal commands and reactions for handovers, as well as the annotation of the activity and handover tasks.
It is worth noting that although we do not use the robot participants' full-body motion information in our experiments (Section 4), yet we release them as additional annotations for the community to work on.

\begin{figure*}[t!]
    \centering
    \includegraphics[width=\linewidth]{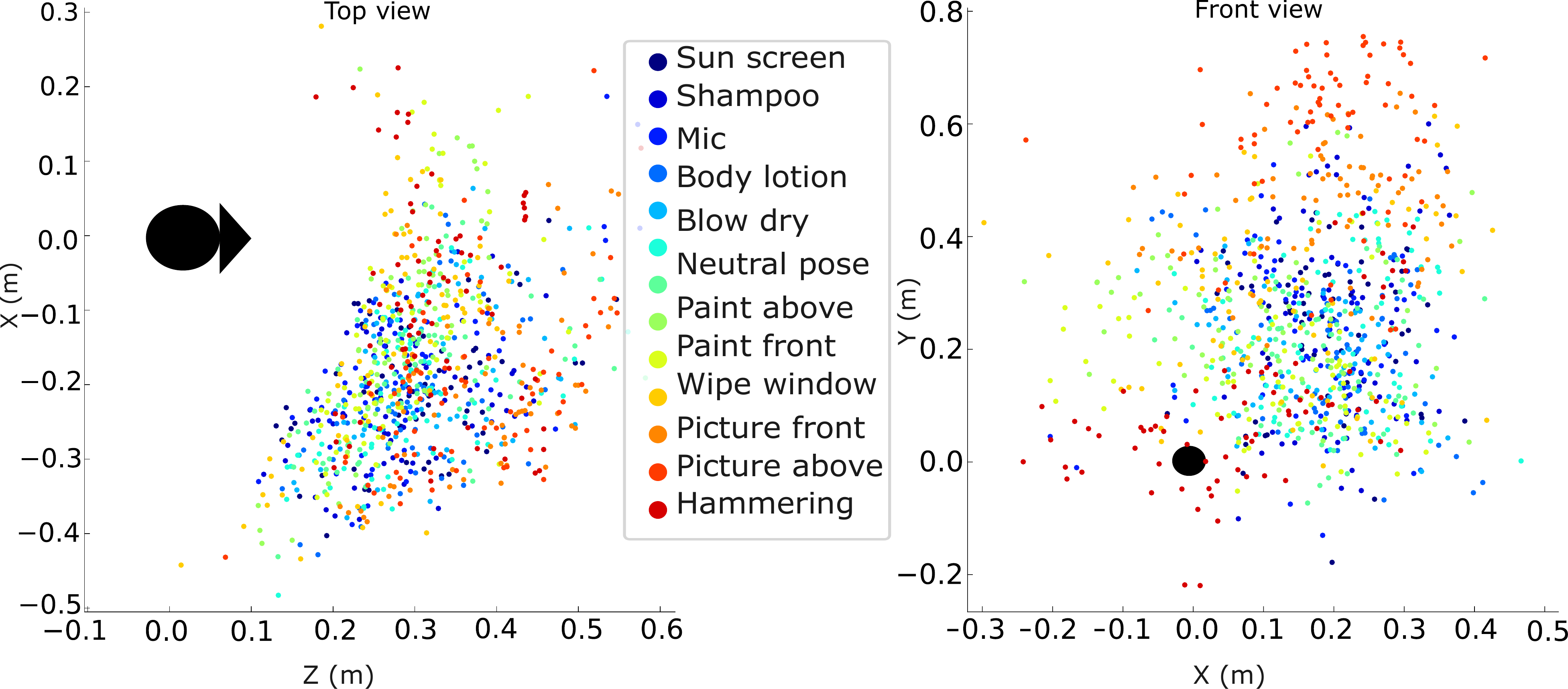}
    \caption{Distribution of the locations where the object was handed over between participants. The points are presented in the user's hips coordinate system. (left) shows the distribution from the top view (head at origin, facing towards right) and (right) from the front view (hip at origin, user facing inwards the plane).}
    \label{fig:stats:pot}
\end{figure*}

\begin{figure*}[t!]
    \centering
    \includegraphics[width=\linewidth]{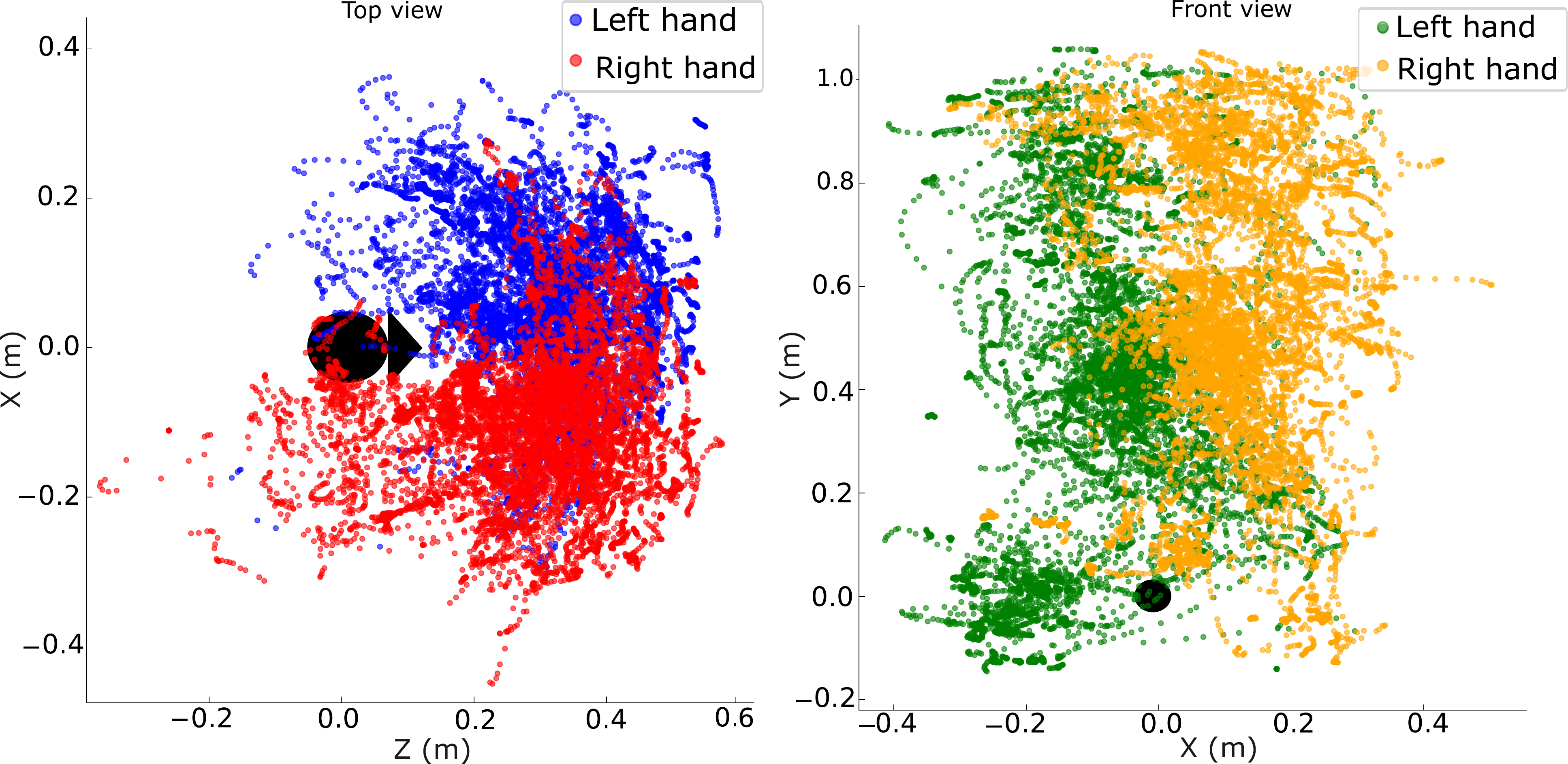}
    \caption{Distribution of the palm over the entire dataset for performing the 12+1 activities. The points are presented in the user's hips coordinate system. (left) shows the top-view
(head at origin, facing towards right), (right) the frontal view (hip at origin, user facing inwards the plane). The color encodes the left and right hands. The unit is meters. 
}
    \label{fig:stats:activity}
\end{figure*}

\subsection{Data Analysis}
We conducted an initial analysis of the dataset to identify significant patterns in the motion data. The average duration of a handover across all activities was $2.244 \pm 0.854$ seconds. 
Another notable aspect of our dataset is where exactly the object was handed over between participants depending on the activity.
\autoref{fig:stats:pot} visualizes the distribution of handover locations in the user's hip coordinate system, color-coded for each activity. The distribution of handovers from the top view (see \autoref{fig:stats:pot} left) shows that the distribution mainly extends to a hemispherical region of up to approx. 0.5~m to the primary user's front and approx. 0.45~m to their right side. Interestingly, it exhibits a distinct skew towards the primary participant's right side, influenced by the positioning of the robot participant on this side. 
The front view (\autoref{fig:stats:pot} right) shows that handovers were primarily performed in an area ranging from hip-level to approx. 0.5 m above hip level, while some handovers, primarily for activities performed at the head level, extend up to approx. 0.8 m above hip level.

Moreover, we were interested in the distribution of motion across all activities because we hypothesized that the varied conditions under which the handovers were performed would result in a wide range of motion patterns, reflecting the flexibility and adaptability of human motor behavior in response to different task demands.
As depicted in \autoref{fig:stats:activity}, the distribution of the primary user's palm position with regards to the user's hip joint throughout the activities reveals extensive coverage across the entire space in front of the user. 
The top view (\autoref{fig:stats:activity} left) and frontal view ( \autoref{fig:stats:activity} right) both show that the primary user utilized a broad range of motion from left to right and head to hip, encompassing nearly all reachable areas, which supports our hypothesis of a large distribution of motion. 
\section{Validating the Dataset with Models}
This section showcases our dataset's usability for generating handover trajectories and predicting key handover characteristics through the following task settings:

\begin{enumerate}
    \item First, we show that our dataset enables training of generative models which \textbf{synthesize handover trajectories} of an SRL in a human-like manner. 
    To this end, we train a conditional variational auto-encoder on the 
    complete handover trajectories of the robotic participant, conditioned on the full-body motion of the primary participant. 
    In effect, this task learns \emph{how} the robotic arm should move.
    \item Secondly, we show that with our data, we can train a model that predicts the \textbf{locations of handover} aligned with the actual handover locations. We train a conditional variational auto-encoder to predict the potential handover position and orientation at any given time on the trajectory. This task informs \emph{where} the robotic arm should move to. 
    \item Lastly, we validate that our dataset contains the vital information for training a binary classifier that predicts \textbf{when a handover occurs} by only observing the primary user's motions. This informs \emph{when} the robot should start to move for a handover. 
\end{enumerate}




For all the tasks described above, we provide details on the data processing steps, the model, and the training process and report technical evaluation results, where the model predictions are compared against the testing data gathered from human participants. We also identify the most influential joints to make training and system deployment more efficient. \\
\subsection{Generating the Trajectory of a Handover} \label{subsection:howTo}

%
%
%
A key aspect that we aim to demonstrate in our dataset is allowing models to generate human-like handover trajectories, which is a challenging task due to the highly variable and high dimensional nature of the human body's motion space. 
Specifically, our goal is to generate motion trajectories from the starting point to the handover position, dynamically accounting for the posture changes of a user during the handover. \\

\subsubsection{Data Processing}
\paragraph{Motion representation.} We define human motion as time series data of sequential human body poses with timeframe $T$. 
At any given timestamp $t$, our dataset contains the positional and rotational data for all joints of both the primary human participant and the robot participant. 
The data processing follows standard methods commonly employed in motion generative and predictive models (e.g., \cite{MVAE}). 
The joint’s position ($j_p$) is represented in the rigged character's root coordinates. Each joint’s rotation ($j_r$) is represented in its local Euler angles. 
%
We normalize the poses by translating and rotating such that the root joint (hip) is positioned at the origin of the world coordinate system and skeletons are oriented uniformly in the same direction.
Finally, to maintain a continuous representation of joint rotations, we project the 3D rotational data into a continuous 6D space ($j_r \in \mathbb{R}^6$), a widely used technique \cite{6d_continuous_representation}.

The pose of the primary participant at timestamp $t$ is expressed as a tuple of joint positions and local rotations, $h^t = [j_p^t, j_r^t]$. 
%
%
%
%
%
%
%
The human motion between timestamp $t_1$ and $t_2$ then is defined as the sequence of poses between $t_1$ and $t_2$: 
$h^{[t_1:t_2]} = [h^{t_1}:h^{t_2}]$\\
Similarly, we define the robot user's pose as $r^t = [EE_p^t, EE_r^t]$ where $EE_p$ and $EE_r$ are the end-effector (right hand of the robot participant) position and rotation respectively. 
%
The robot motion is then $r^{[t_1:t_2]} = [r^{t_1}:r^{t_2}]$. 
\paragraph{Input and output data.}
There are three inputs to our model: the primary participant's full-body motion, the robot participant's end-effector motion, and the handover state ("handing over", "taking back," or "idle"). 
The motions of the primary and robot participants are symmetric time windows looking back $T$ timestamps into the past and $T$ timestamps into the future. 
We can describe such input data of the primary participant as $h^{[t-T:t+T]}$ and the robot participant as $r^{[t-T:t+T]}$. 
We set $T=25$, corresponding to a duration of one second. 
The output of our model is the generated 3DoF next position of the robot end-effector mapped to the robot participant's right hand from the \dataset. 
%
%
%
%

%
\paragraph{Train and test data separation.}
Our data comes from human subjects, where each participant's data is likely to contain unique patterns or behaviors. 
%
%
We perform a train-test split on the participant level to ensure the test data is completely unused for training. Specifically, we randomly select two participant pairs (i.e., 4 participants' data) as the test set, which are not involved in model training.
The remaining 8 participants' data is utilized for training, and not involved in testing at all.
This data separation is consistent throughout all the following models.
\subsubsection{Model} 
Conditional Variational Autoencoders (CVAE) \cite{sohn2015CVAE} are one of the most popular architectures widely used for generating motions or poses \cite{starke2022genmotion-vae1, diomataris2024wandr-genmotion-vae2, MVAE, ghosh2023imos-genmotion-cvae1, rempe2021humor-cvae2}, with an encoder-decoder structure. 
The encoder compresses high-dimensional pose data into a smooth, continuous latent space ($z$), while the decoder generates the next pose based on this representation and conditional inputs. 
By modeling the probabilistic distribution of the data, CVAEs can generate diverse and realistic motion sequences through sampling. 
Among CVAE-based models, our approach builds upon the Motion Variational Autoencoder (MVAE) \cite{MVAE}. 
We chose MVAE as the basis for our model because of its demonstrated capabilities in autoregressive motion generation making its output robust against the highly variable motion inputs. 
MVAE utilizes a Mixture of Experts (MoE) architecture in the decoder, refining predictions by incorporating multiple specialized networks for different motion aspects.

\begin{figure}[t!]
    \centering
    \includegraphics[width=1.0\linewidth]{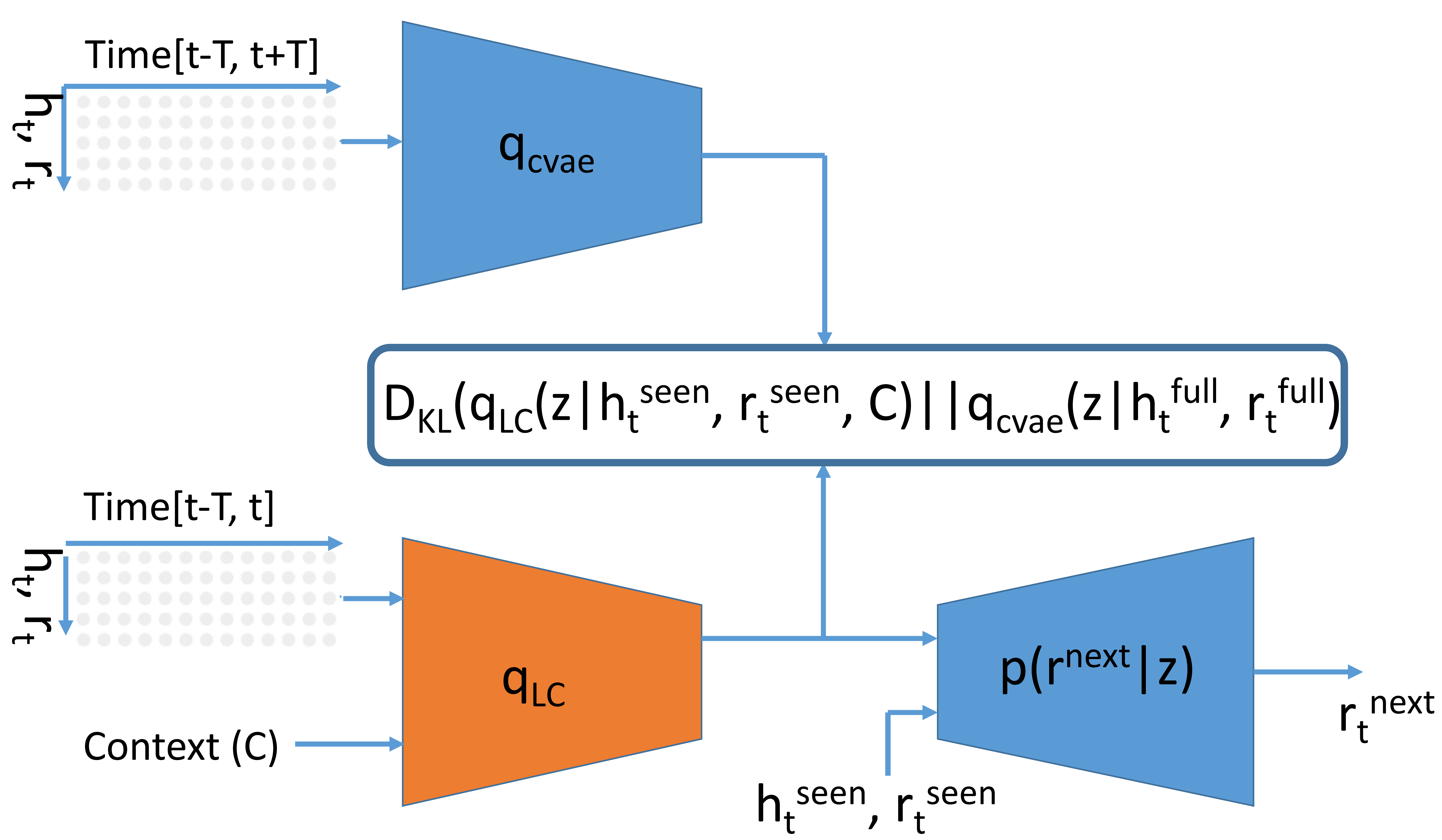}
    \caption{Overview of our proposed model architecture for generating the trajectory of a handover based on the motion dynamics encoded into the model's latent space.}
    \label{fig:howTo-pipeline}
\end{figure}

Building on MVAE, we propose \textbf{SVAE} (\textbf{CVAE} for \textbf{S}RL), a model specifically designed for SRL's motions in handover processes. The model architecture is illustrated in Figure~\ref{fig:howTo-pipeline}. 
While MVAE and SVAE share the same foundational framework, SVAE addresses two novel challenges: generating SRL motion by considering both the user's and the robot's movements and adapting to varying handover states, which MVAE does not account for.
A key enhancement in SVAE is the integration of attention mechanisms across its components enabling the model to better capture the relevant temporal and spatial aspects of the input motion data. 
The \textit{encoder} in SVAE, similar to MVAE, compresses high-dimensional data into a latent space, but it processes motion data over a time window that includes current, past, and future timesteps, unlike MVAE, which only handles current and past data. 
The \textit{decoder} generates the next pose by sampling from the latent space, but in contrast to MVAE, our model is conditioned on both the robot's and human's observed motions. 
SVAE keeps the \textit{Mixture of Experts (MoE)} architecture for the decoder, utilizing six expert networks and a gating mechanism. 
Additionally, our model introduces a \textit{latent controller (LC)}, which aligns the latent space with the specific handover state ("handing in," "taking away," or "idle"). 
This context-aware layer provides enhanced control, allowing SVAE to manage the timing and variability of human-robot interactions during handovers, an essential feature not present in MVAE. 
The latent controller also incorporates an attention mechanism to adaptively align the input motions and handover state with the latent representation learned by the encoder.

Here, we provide the loss function based on the Evidence Lower Bound (ELBO) to train the SVAE :

\begin{align*}
ELBO_t^r &= \mathbb{E}_{q{LC}}\left[\log p_\theta(r^{next}|z, h^{seen}, r^{seen})\right] \\
&\quad - \beta \, \text{KL}\left(q_{LC}(z|h_t^{\text{seen}}, r_t^{\text{seen}}, C) \, || \, p(z|h_t^{\text{full}}, r_t^{\text{full}})\right)
\end{align*}

\noindent where \( t \) represents the current timestamp. \( h\) and \( r\) refer to the primary participant and robot participant, respectively. The loss function has two main components. 
The first term is the reconstruction loss \( \mathbb{E}_{q_{LC}}\left[\log p_\theta(r^{next}|z, h^{seen}, r^{seen})\right] \). It measures how well the decoder can predict the next SRL pose \( r^{next} \). 
\( z \) is the latent variable encoding the motion information, \( h^{seen} \) represents the observed human motion, and \( r^{seen} \) represents the observed SRL motion up to the current timestamp. This term ensures that the model generates accurate and contextually appropriate motions. 
The second term, $KL$, is the Kullback-Leibler (KL) divergence, which regularizes the latent space by aligning the learned posterior distribution \( q_{LC}(z|h_t^{\text{seen}}, r_t^{\text{seen}}, C) \) with a prior distribution \( p(z|h_t^{\text{full}}, r_t^{\text{full}}) \). Here, \( h_t^{\text{full}} \) and \( r_t^{\text{full}} \) represent the full human and SRL motion data across the time window, and \( C \) denotes the handover state ("handing in," "taking away," or "idle"). 
The parameter \( \beta \) controls the trade-off between the two components in the ELBO. Balancing the model's ability to reconstruct accurate motions and maintaining a well-structured latent space that generalizes effectively, we used $\beta=0.1$ in our setting. 

\subsubsection{Training}

Our training process is divided into two main stages. 
In the first stage, we train the SVAE model for 140 epochs. 
Following that, we use the trained SVAE to train the complete pipeline, which includes both the SVAE and the LC encoder, for 250 epochs. 
We begin with training the SVAE model for 10 epochs using only the reconstruction loss, after which we introduce KL divergence loss into the loss function. 
\[
\mathcal{L}_{\text{SVAE}} = \mathcal{L}_{rec} + \beta \cdot \mathcal{L}_{\text{KL}(z_{\text{SVAE}}, \mathcal{N}(0,I))}
\]
We employ the adaptive moment estimation (ADAM) optimizer, with a learning rate that decays from $10^{-4}$ to $10^{-7}$, starting to decay at the 50\textsuperscript{th} epoch and continuing throughout the training. 

In the second stage, we align the output of the LC encoder with the learned latent space of the SVAE while freezing the weights of the encoder in SVAE. 
For the first 50 epochs, we minimize a loss based on the KL divergence between the learned latent space $z_{SVAE}$ and the latent of the LC encoder $z_{LC}$. 
After that, the reconstruction error is incorporated into the loss, and training continues for an additional 50 epochs. 
The loss function for the training process is:
\[
\mathcal{L}_{LC,SVAE} = \mathcal{L}_{rec} + \beta \cdot \mathcal{L}_{\text{KL}(z_{\text{SVAE}}, z_{\text{LC}})}
\]
We trained our model using reconstruction loss based on the $\mathbf{l_2}$ distance of the 3DoF of the robot’s end-effector ($\hat{y}$), which corresponds to the right hand of the SRL participant from the dataset ($y$). 

We apply scheduled sampling during training, where the model's output is fed back as input for autoregressive generation over $l=10$ consecutive steps. 
The probability of using autoregression, $p$, increases from 0 to 1 over 50 epochs, after which the model is fully autoregressive ($p=1$). 
The ADAM optimizer is used again in this stage, with learning rate settings similar to the first stage.

\subsubsection{Testing Results} 

We generate trajectories in two forms. 
One is non-autoregressive: we take the seen motion sequence of the test dataset to generate a single next position ($\hat{y}_t$) and evaluate it against the ground truth position in the test data ($y_t$). 
The other is autoregressive: we provide the model with an initial starting position, which then continuously generates the next positions based on the very previously generated 25 robot positions. 

To quantify the quality of the generated trajectories, we report the Mean Absolute Error (MAE) of pairwise comparison between the generated and the ground-truth trajectories: $\text{MAE} = \frac{1}{\eta} \sum_{j=1}^{\eta} \frac{1}{N} \sum_{i=1}^{N} |y_t - \hat{y}_t|$, where N is the length of the trajectories normalized by the number of trajectories $\eta$. 

Table~\ref{table:howTo_test_results} shows the MAE errors. Our model generates handover trajectories with MAEs ranging between 2.10--2.71~cm in the non-autoregressive setting. With autoregression, MAEs range between 10.42--23.85~cm.
We observe that our model shows strong performance in the pairwise comparisons with the ground-truth in the non-autoregressive setting. The error increases in the autoregressive setting, which is expected as errors accumulate. It is important to highlight a key distinction between our motion generation task and previous related work (e.g., \cite{MVAE}). In most prior approaches, motion generation is based on observations of the same actor. In contrast, our model generates the motions of one actor (the robot user) based on the observations of a different actor (the primary user). This fundamental difference introduces additional complexity. Both actors interact in a continuous real-time feedback loop, where the motion of one most likely directly influences the motion of the other. This dynamic interplay cannot be fully accounted for in the autoregressive setting. It is to be assumed that in interactive real-world deployments, the error will be lower, as the primary user would adapt their motion trajectory to the robot's trajectory.   
As this is a novel research question, future works should investigate more advanced models to further mitigate the error. One potential idea is to employ reasoning models, which infer the primary users' intents and use that to condition or correct the robot's motions. 

For a qualitative impression of the generated trajectories, we refer the reader to the Video Figure where we show several representative trajectories generated for different activities. Results show that our model generates plausible trajectories that are in keeping with the key characteristics of naturalistic human handover motion in close peripersonal space. 

To quantify the individual joints' importance for generating motions, we employ a gradient-based sensitivity analysis \cite{kovacs2019gradient-based-sensitivity1,most2024sensitivity2}. A higher gradient indicates this input feature has a higher influence on the output. The result is shown in Table~\ref{table:howTo_joint_importance_rank}. Our findings suggest that, on average, the positions of the "left shoulder", "right elbow" and "left hand" joints are most influential, while the rotations of the "Face", "right elbow" and "right hand" offer the most decisive rotation information for generating the handover trajectory. The neck's position and the back's rotation are the least decisive features.

\def\mic{Mount a mic}
\def\sunscreen{Apply sunscreen to face}
\def\lotion{Apply body lotion to chest}
\def\shampooing{Shampoo hair}
\def\washing{Wash torso with washcloth}
\def\drying{Blow dry hair}
\def\pictureLow{Straighten a pic (low)}
\def\pictureHigh{Straighten a picture (high)}
\def\hammering{Hammer a nail}
\def\cleaning{Clean a window}
\def\paintingLow{Paint the wall (low)}
\def\paintingHigh{Paint the wall (high)}
\def\neutral{Neutral pose}

\begin{table}[t!]
    \centering
    \begin{tabular}{p{3.5cm}|p{1.78cm} p{1.78cm}}
    \multirow{2}{*}{\textbf{Activity}} & \textbf{MAE W/O} & \textbf{MAE W/}\\ 
    & \textbf{autoreg. ($cm$)} & \textbf{autoreg. ($cm$)}
    \\ \hline
    \hline
    \mic & $2.55 \pm 2.61$ & $15.22 \pm 16.52$ \\ 
    \rowcolor{LightGray}
    \sunscreen & $2.70 \pm 3.10$ & $14.13 \pm 15.95$ \\ 
    \lotion & $2.60 \pm 2.34$ & $15.09 \pm 16.67$ \\ 
    \rowcolor{LightGray}
    \shampooing & $2.57 \pm 2.25$ & $12.77 \pm 15.73$ \\ 
    \washing & $02.55 \pm 2.33$ & $15.03 \pm 14.51$ \\ 
    \rowcolor{LightGray}
    \drying & $2.55 \pm 2.32$ & $15.77 \pm 15.58$ \\ 
    \pictureLow & $2.34 \pm 2.03$ & $22.28 \pm 24.55$ \\ 
    \rowcolor{LightGray}
    \pictureHigh & $2.60 \pm 2.84$ & $23.85 \pm 28.23$ \\ 
    \hammering & $2.10 \pm 1.80$ & $10.42 \pm 8.71$ \\ 
    \rowcolor{LightGray}    
    \cleaning & $2.71 \pm 2.50$ & $14.79 \pm 15.85$ \\ 
    \paintingLow & $2.51 \pm 2.43$ & $10.72 \pm 10.54$ \\ 
    \rowcolor{LightGray}    
    \paintingHigh & $2.59 \pm 2.15$ & $14.15 \pm 16.59$ \\ 
    \neutral & $2.60 \pm 2.50$ & $12.48 \pm 13.88$ \\ 
    \end{tabular}
    \caption{Results for generating the trajectory of a handover. Mean absolute error in meters (m) is reported.}
    \label{table:howTo_test_results}
\end{table}

\begin{table}[t!]
    \centering
    \begin{tabular}{p{0.7cm}|p{2.3cm} p{2cm}}
    \textbf{JIR} & \textbf{Joint Position} & \textbf{Joint Rotation}\\ \hline \hline
    \textbf{1} & Left shoulder & Face \\ 
    \rowcolor{LightGray}
    \textbf{2} & Right elbow & Right elbow \\ 
    \textbf{3} & Left hand & Right hand \\ 
    \rowcolor{LightGray}
    \textbf{4} & Chest notch & Left hand \\ 
    \textbf{5} & Head & Right wrist \\ 
    \rowcolor{LightGray}
    \textbf{6} & Right shoulder & Upper back \\ 
    \textbf{7} & Right hand & Left clavicle \\ 
    \rowcolor{LightGray}
    \textbf{8} & Back & Head \\ 
    \textbf{9} & Right clavicle & Right clavicle \\ 
    \rowcolor{LightGray}
    \textbf{10} & Face & Left elbow \\ 
    \textbf{11} & Upper back & Left wrist \\ 
    \rowcolor{LightGray}
    \textbf{12} & Write wrist & Right shoulder \\ 
    \textbf{13} & Left clavicle & Neck \\ 
    \rowcolor{LightGray}
    \textbf{14} & Left elbow & Chest notch \\ 
    \textbf{15} & Left wrist & Left shoulder \\ 
    \rowcolor{LightGray}
    \textbf{16} & Neck & Back \\ 
    \end{tabular}
    \caption{Joint Importance Rank (JIR) for handover trajectory generation, indicating the relevance of an individual joint's position and rotation generating motions.}
    \label{table:howTo_joint_importance_rank}
\end{table}


\subsection{Generating the Region of Transfer} \label{subsection:wehreTo}

%


The 3D location where the object is a crucial characteristic in handover motions, and so is the rotation of hands at the time of handover. 
Such information regarding the position and rotation of the handover is also known as the ``region of Transfer'' or ROT \cite{wiederhold2024hoh}. 
Given that this ROT is the final product of a full trajectory, ideally, it can be predicted by the observed segments of the trajectory. 
In this subsection, we show that our dataset contains detailed motions that enable ROT prediction. 
\subsubsection{Data Processing.} 
Following \citet{wiederhold2024hoh}, we define the handover coordinate (location) as the midpoint and the orientation as the direction of the axis that passes through the palms of the primary participant and robot participant at the time of handover. 
At each timestamp $t$ during the handover, the input to our model is the primary user's motion $h^{[t-T:t]}$ and the robot user's motion $r^{[t-T,t]}$ over the past time period T=25 (1 sec). 

We acquire the ground truth handover coordinate and rotation as the 3D mid-point of user-SRL hands and the 3D vector pointing from the giver's hand to the receiver's hand that shows the orientation of the user-SRL hands at the transfer time stamp. 
The output of our model is the 6 DoF-generated ROT for the current motion of the user-SRL. 
%
%

\subsubsection{Model} 
The generation of the RoT is very aligned with generating the trajectories in the previous subsection, as both are taking the primary user's motions as input, however, the main difference is the timing of the output. 
While the trajectory generation relies on the autoregressive generation of the next positions sequentially, the RoT generation does not have this constraint. 
%
Therefore, we employ a conditional variational autoencoder (CVAE) \cite{sohn2015CVAE} for this task because its capability in pose generation has been demonstrated in previous work \cite{ghosh2023imos-genmotion-cvae1, rempe2021humor-cvae2}. 
The encoder $\phi$ in our CVAE encodes the input motions into the latent value $z$. 
The decoder $\theta$ in our model samples from the latent distribution $z$ and, conditioned on the human and the SRL motions, generates the 6 DoF ROT. 

\subsubsection{Training}
We train the pipeline for 250 epochs with an adaptive moment estimation (ADAM) optimizer, with a decaying learning rate from $10^{-4}$ to $10^{-7}$. 
The model is trained with $l_2$ distance for position and orientation of the RoT. 

\[
\mathcal{L} = \frac{1}{2} \left( \|\mathbf{\hat{p}} - \mathbf{p}\|_2^2 + \|\mathbf{\hat{q}} - \mathbf{q}\|_2^2 \right)
\]
where $\hat{p}$ and $p$ are the generated and ground truth 3DoF positions respectively, and $\hat{q}$ and $q$ are the generated and ground truth 3 DoF orientations. 

\subsubsection{Testing Results} We examine the performance of the model in predicting the 6 DoF features of the RoT at each time stamp during the handover process by observing the motions from the past 1 second. 
Table~\ref{table:whereTo_test_results} reports the mean absolute error (MAE) for the 3 DoF position and the mean Euler angle error (MEAE) for the orientation of RoT. 
The results show that the model achieves MAEs that range between 4.02cm and 8.04cm, while the achieved MEAE is between 0.0002 and 0.004 radians. These relatively low errors indicate that our dataset captures sufficient data to allow for predicting the Region of Transfer.

Furthermore, we investigate the importance levels of individual joints of the primary user's and the SRL's motions for predicting the RoT information. The results are shown in Table~\ref{table:joint_importance_rank_whereTo}. The left elbow's position and rotation are reported to be the most influential feature impacting the whereabouts of the RoT. 

\begin{table}[t!]
    \centering
    \begin{tabular}{p{3.5cm}|p{1.4cm} p{1.9cm}}
    \textbf{Activity} & \textbf{MAE ($cm$)} & \textbf{MEAE ($rad$)} \\ \hline \hline
    \mic & 6.24 $\pm$ 2.82 & 0.0207 $\pm$  0.0142 \\ 
    \rowcolor{LightGray}            
    \sunscreen & 7.37 $\pm$ 3.67 &  0.0268 $\pm$  0.0196 \\ 
    \lotion & 6.01 $\pm$ 2.85 & 0.0211 $\pm$  0.0186 \\ 
    \rowcolor{LightGray}
    \shampooing & 8.04 $\pm$ 3.14 & 0.0165 $\pm$  0.0121 \\ 
    \washing & 6.71 $\pm$ 3.12 & 0.0221 $\pm$  0.0125 \\ 
    \rowcolor{LightGray}
    \drying & 7.45 $\pm$ 3.01 & 0.0240 $\pm$  0.0176 \\ 
    \pictureLow & 4.88 $\pm$ 2.45 & 0.0076 $\pm$  0.0054 \\ 
    \rowcolor{LightGray}
    \pictureHigh & 5.69 $\pm$ 3.41 & 0.0092 $\pm$  0.0088 \\ 
    \hammering & 4.02 $\pm$ 2.49 & 0.0099 $\pm$  0.0067 \\ 
    \rowcolor{LightGray}
    \cleaning & 6.13 $\pm$ 3.26 & 0.0159 $\pm$  0.0107 \\ 
    \paintingLow & 4.20 $\pm$ 2.13 & 0.0132 $\pm$  0.0097 \\ 
    \rowcolor{LightGray}
    \paintingHigh & 6.41 $\pm$ 3.83 & 0.0179 $\pm$  0.0124 \\ 
    \neutral & 4.72 $\pm$ 2.57 & 0.0102 $\pm$  0.0100 \\ 
    \end{tabular}
    \caption{Test results for generating the region of transfer: mean absolute error of the generated positions (left) and rotation angles (right).}
    \label{table:whereTo_test_results}
\end{table}

\begin{table}[t!]
    \centering
    \begin{tabular}{p{0.7cm}|p{2.3cm} p{2cm}}
    \textbf{JIR} & \textbf{Joint Position} & \textbf{Joint Rotation}\\ \hline \hline
    \textbf{1}  & Left elbow & Left elbow \\ 
    \rowcolor{LightGray}
    \textbf{2}  & Chest notch & Chest notch \\ 
    \textbf{3}  & Face & Left shoulder \\ 
    \rowcolor{LightGray}
    \textbf{4}  & Head & Right shoulder \\ 
    \textbf{5}  & Right wrist & Right elbow \\ 
    \rowcolor{LightGray}
    \textbf{6}  & Right clavicle & Right wrist \\ 
    \textbf{7}  & Left wrist & Neck \\ 
    \rowcolor{LightGray}
    \textbf{8}  & Left clavicle & Upper back \\ 
    \textbf{9}  & Left hand & Face \\ 
    \rowcolor{LightGray}
    \textbf{10} & Right elbow & Left wrist \\ 
    \textbf{11} & Left shoulder & Left hand \\ 
    \rowcolor{LightGray}
    \textbf{12} & Right hand & Right clavicle \\ 
    \textbf{13} & Back & Head \\ 
    \rowcolor{LightGray}
    \textbf{14} & Upper back & Left clavicle \\ 
    \textbf{15} & Right shoulder & Back \\ 
    \rowcolor{LightGray}
    \textbf{16} & Neck & Right hand \\ 
    \end{tabular}
    \caption{Joint importance rank (JIR) for generation of Region of Transfer, indicating the relevance of an individual joint's position and rotation. }
    \label{table:joint_importance_rank_whereTo}
\end{table}

\subsection{Predicting the Timeframe of Handover}\label{subsection:whenTo}
%
%
%
We demonstrate our dataset's capability to predict the moment the primary user wants to initiate a handover, just from observing the primary user's motions. 
Similar to how humans use implicit cues without verbal expressions in human-to-human handover~\cite{predict-human-intention}, we demonstrate that our dataset encapsulates such implicit cues and thus allows for training a model for prediction. We define this problem as a binary classification problem, where the model is trained to predict whether a handover is currently ongoing or not. 

\subsubsection{Data Processing} 
As "handover", we define the sequence of frames that begins when the robot user starts moving, and that ends when the robot user’s hand has returned to its resting position after the object has been handed over. Our dataset comprises ground-truth annotation with a binary variable $y$ that indicates for each frame whether it belongs to a "handover" or not. 

%
%
At each timestamp, $t$, the input to the model consists of a sliding window of the user’s motion data over a time window of length $T=25$ (1 sec) previous to the current timestamp $h^{[t-T:t]}$. 
The model's output at each timestamp $t$ is a continuous float value representing the likelihood of a handover being in progress at $t$. We use thresholding to convert this likelihood into a binary classification result: any value greater than 0.6 is considered a "handover". 
\subsubsection{Model} We employ a model composed of a 3-layered fully connected neural network with 128 nodes in each layer. For the activation functions, the first two layers are followed by the ELU function, and the last layer is followed by the Sigmoid function after the last linear layer, to ensure the output is bounded to $[0,1]$. 
%

%
\subsubsection{Training} We trained our model on all instances of handover in the train data, regardless of how the participants had communicated their handover intent. 
We used the same optimizer type (ADAM) and decaying learning rate ($10^{-4}$ to $10^{-7}$ as in our previous experiments. 
We trained the model for 500 epochs with the binary cross-entropy loss function. 
\begin{equation*}
\mathcal{L} = - \frac{1}{N} \sum_{t=1}^{N} \left[ y^t \log(\hat{y}^t) + (1 - y^t) \log(1 - \hat{y}^t) \right]
\end{equation*}

\subsubsection{Testing Results}

Results of classification accuracy are detailed in Table~\ref{table:accuracy_results}. Across all activities, the model achieves an accuracy of 84.3\%. They were highest (100\%) for washing the torso with a washcloth activity and lowest (66.7\%) for the hammering a nail activity. We have also analyzed whether the parameters of the user's activity (height, distance from the body, motion range, see Section~\ref{sec:Activities}) have an influence on how accurately an intended handover can be identified. The model has achieved the maximum accuracy for the activities performed close to the body (90.3\%), and a somewhat lower accuracy for activities away from the body (79.5\%). The accuracy of the model is also highest for activities comprising a large motion range (88.5\%) and small motion range (85.0\%), and slightly lower for activities with a medium motion range (79.2\%). The height of activities does not affect classification accuracy ($\approx$84\% for both head and torso levels).
Table~\ref{table:joint_importance_rank_whenTo} shows the result of the Joint Importance Rank (JIR) analysis. It reveals that the position of the left wrist and the rotation of the neck are the most impactful features in the primary user's motion features that can convey that a handover is happening.

\begin{table}[t!]
    \centering
    \begin{tabular}{p{3.7cm}|p{1.4cm}}
    \textbf{Activity} & \textbf{Accuracy} \\ \hline \hline
    \mic & 91.7\% \\ 
    \rowcolor{LightGray}
    \sunscreen & 91.7\% \\ 
    \lotion & 91.7\% \\ 
    \rowcolor{LightGray}
    \shampooing & 83.3\% \\ 
    \washing & 100\% \\ 
    \rowcolor{LightGray}
    \drying & 83.3\% \\ 
    \pictureLow & 75.0\% \\ 
    \rowcolor{LightGray}  
    \pictureHigh & 83.3\% \\ 
    \hammering & 66.7\% \\ 
    \rowcolor{LightGray}
    \cleaning & 75.0\% \\ 
    \paintingLow & 81.3\% \\ 
    \rowcolor{LightGray}
    \paintingHigh & 91.7\% \\ 
    \neutral & 83.3\% \\ 
    \rowcolor{LightGray}
    \textbf{Overall} & \textbf{84.4\%} \\ 
    \end{tabular}
    \caption{Classification accuracy for predicting the time frame of handover.}
    \label{table:accuracy_results}
\end{table}

\begin{table}[t!]
    \centering
    \begin{tabular}{p{0.7cm}|p{2.3cm} p{2cm}}
    \textbf{JIR} & \textbf{Joint Position} & \textbf{Joint Rotation}\\ \hline \hline
    \textbf{1}  & Left wrist & Neck \\ 
    \rowcolor{LightGray}
    \textbf{2}  & Chest notch & Right shoulder \\ 
    \textbf{3}  & Head & Right elbow \\ 
    \rowcolor{LightGray}
    \textbf{4}  & Chest notch & Left elbow \\ 
    \textbf{5}  & Upper back & Face \\ 
    \rowcolor{LightGray}
    \textbf{6}  & Back & Back \\ 
    \textbf{7}  & Right hand & Left shoulder \\ 
    \rowcolor{LightGray}
    \textbf{8}  & Right elbow & Right wrist \\ 
    \textbf{9}  & Face & Left wrist \\ 
    \rowcolor{LightGray}
    \textbf{10} & Right wrist & Right arm \\ 
    \textbf{11} & Left arm & Left arm \\ 
    \rowcolor{LightGray}
    \textbf{12} & Left elbow & Chest notch \\ 
    \textbf{13} & Neck & Upper back \\ 
    \rowcolor{LightGray}
    \textbf{14} & Right shoulder & Head \\ 
    \textbf{15} & Left shoulder & Right hand \\ 
    \rowcolor{LightGray}
    \textbf{16} & Right arm & Left hand \\ 
    \end{tabular}
    \caption{Joint Importance Rank (JIR) for predicting the timeframe of handover, indicating the relevance of an individual joint’s position and rotation.}
    \label{table:joint_importance_rank_whenTo}
\end{table}

\section{User Study: Validating the Perceived Quality of the Handover Interaction}

We conducted a user study to compare the perceived quality of overall handover interactions generated by our data-driven method, trained on the 3HANDS dataset, with an established baseline method for performing handovers with an SRL~\cite{AjoudaniBiDirecontalHandover}.
To focus on validating the efficacy of our dataset and its capability to enable generative models while minimizing confounding variables potentially introduced by a specific hardware implementation, we carried out the user study in a virtual reality (VR) environment.

\subsection{Experiment Design}
The study employed a within-subject design. 
The participants were asked to perform handover interactions with a virtual SRL, where the SRL's motions are generated by the baseline and 3HANDS data-driven methods. 
To constrain the overall study duration to an hour and still allow for two repetitions and a wide range of different motions, we selected 6 out of 13 total activities.
%
The handover approaches were counterbalanced with a Balanced Latin Square to mitigate order effects.
After experiencing one approach, the participants were asked to respond to eight 7-point Likert questions (see \autoref{fig:likertScaleResults}), focusing on the aspects of naturalness, comfort, physical demand, predictability, timing, smoothness, and appropriateness.
We recruited a total of 10 participants (5 male, 5 female, aged 16-58). The participants received monetary compensation for their participation in the study.

\subsection{Motion Generation Approaches}
We implemented the following two methods:

\subsubsection{Generative Models Trained with the 3HANDS Dataset}
Our approach integrates the \textbf{trajectory generating} SVAE model (\autoref{subsection:howTo}) and the \textbf{handover timeframe predicting} model (\autoref{subsection:whenTo}).
The \textbf{handover timeframe predicting} model continuously monitors the user's motions to determine whether the user is initiating a handover process or engaged in another activity. 
Once the \textbf{handover-timeframe-predicting} model detects that the user has initiated a handover process, the SVAE model then begins generating SRL's motion autoregressively, based on both the current and past motions of the user and the SRL.
The handover is completed when the SRL's end-effector comes within close distance of the user's hand, at which point the SRL halts its motion to finalize the transfer. 
To prevent the object from entering the hand's simulation and causing object-hand collisions, we set the distance threshold to 12 cm, taking into account the object's size of 10 cm.
%
In the following, we refer to our approach as \emph{3HANDS}.


\subsubsection{Baseline Approach}
The most established approach to drive an SRL to complete handover motion is by predicting the user's hand position via extrapolation. 
We chose a baseline implementation from closely related prior work that shares the same SRL-centric setting with our scenario \cite{AjoudaniBiDirecontalHandover}. 
In this approach, pre-defined activation regions in the workspace serve as triggers for the SRL. When the user places their hand within one of these predefined 3D volumes, the SRL recognizes the user's intention to initiate a handover.
Once activated, the SRL relies on a Kalman filter to predict the next 3D position of the user’s hand until the handover is complete.

Using the predicted 3D position, the SRL calculates a trajectory to approach the user’s hand. 
The SRL stops its movement when it reaches a predefined distance (12~cm in our implementation) from the user’s hand, waiting for the object to be transferred.
The SRL in the original paper utilized a 6-degree-of-freedom (6DoF) configuration, with 3DoF dedicated to reaching the goal position and the additional 3DoF used for collision avoidance. However, in our study, the SRL has a 3-degree-of-freedom (3DoF) configuration, focusing solely on the end-effector's position because collision avoidance is not the focus of our study.
In the following, we refer to this approach as \emph{baseline}.

\begin{figure}[t!]
    \centering
    \includegraphics[width=1.0\linewidth]{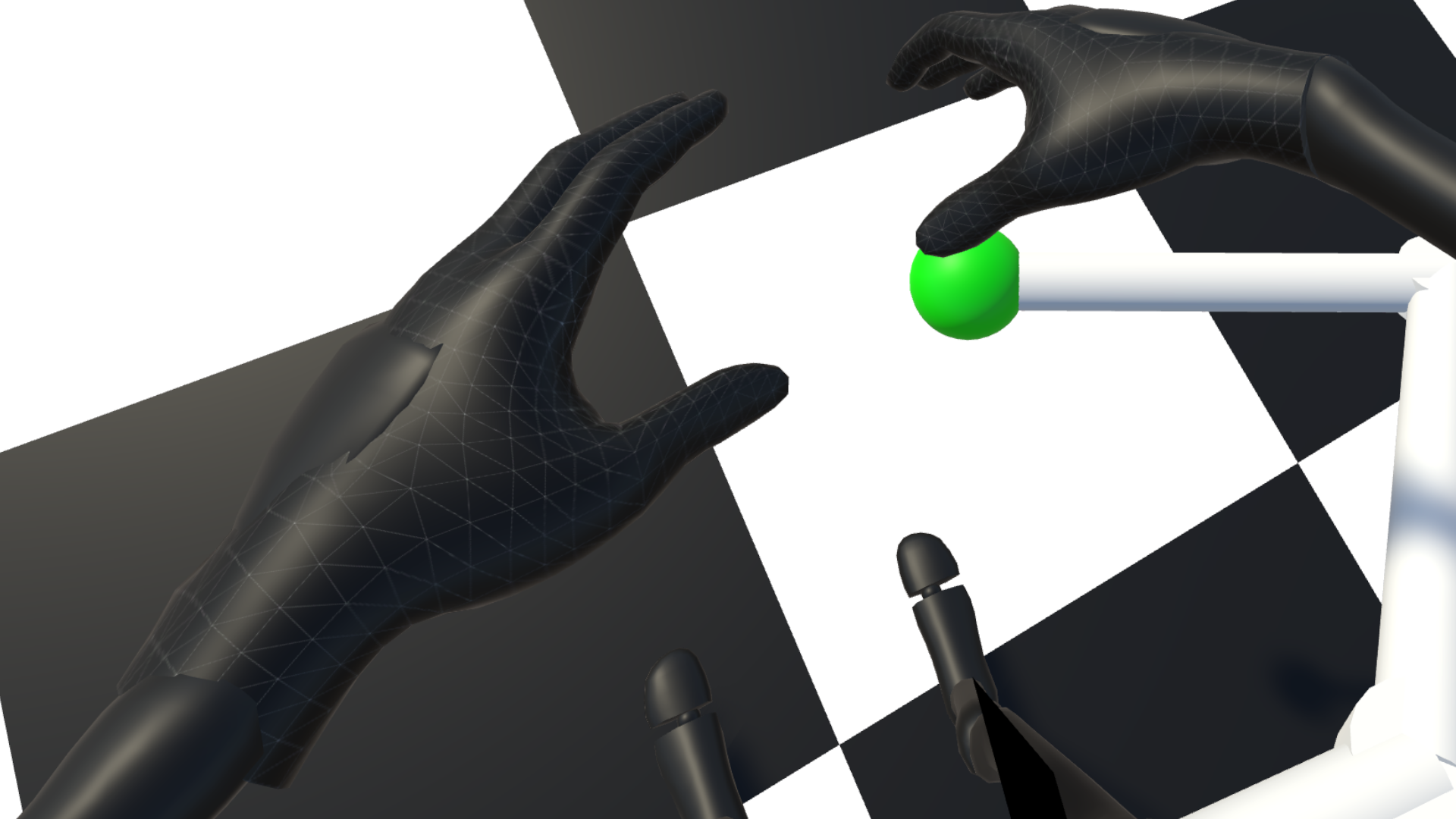}
    \caption{First person view in VR of the handover interaction: Participants performed a task and then instructed the SRL to hand over the green ball either using the 3HANDS or baseline method.}
    \label{fig:studysetup}
\end{figure}

\subsection{Apparatus and Task}
The experimental setup to evaluate the handover approaches was implemented in a virtual reality (VR) environment using Unity3D, run on a Quest Pro VR set. 
The whole experiment was run on Windows 10 with NVIDIA GeForce RTX 4090 GPU. 
The participants observed a humanoid representation of themselves in VR, and an SRL was virtually mounted on their hip (see \autoref{fig:studysetup}). 
The object for the handover was a sphere positioned at the end-effector of the SRL.
During the experiment, the user's and SRL's current poses were transmitted to the model at each time step. 
The model then generated the SRL's subsequent position, both during handover interactions and while the SRL remained idle.
%

\subsection{Procedure}
Participants were first introduced to the study, followed by a tutorial for both approaches. Then, each participant performed 24 handover trials (2 approaches, 6 tasks, 2 repetitions). 
After each trial, they answered the 8 questions.

\subsection{Results and Discussion}
\begin{figure*}
    \centering
    \includegraphics[width=\linewidth]{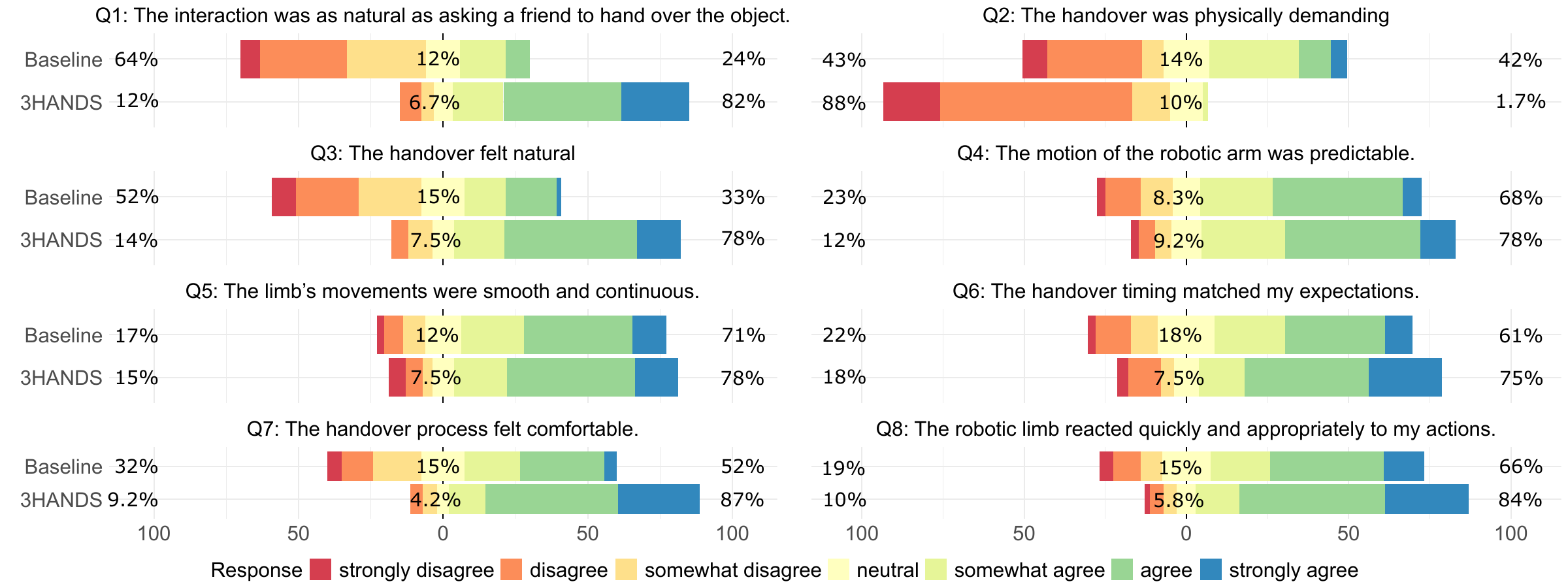}
    \caption{Results of the 7-point Likert scale for qualitative questions comparing the data-driven 3HANDS and baseline approaches.}
    \label{fig:likertScaleResults}
\end{figure*}

We analyzed the Likert ratings using Wilcoxon Signed Rank Tests. 
The results are presented in Figure \ref{fig:likertScaleResults}.

\subsubsection{Perceived Naturalness}
Question 1 (\emph{The interaction was as natural as asking a friend to hand over the object}) and question 3 (\emph{The handover felt natural}) focus on the user's perceived naturalness.
We asked two questions to capture two facets of naturalness: Q1 focuses on interpreting naturalness as interacting with a friendly human, while question 3 leaves more room for a broader interpretation.
We found significant effects for both question 1 (3HANDS median = 6, baseline median = 3, W = 2086, p < .001) and question 3 (3HANDS median = 6, baseline median = 3, W = 3197, p < .001).
The results indicated that 3HANDS allows for higher perceived naturalness.

\subsubsection{Perceived Physical Demand and Comfort}
Question 2 (\emph{The handover was physically demanding}) and question 7 (\emph{The handover process felt comfortable}) are related to the perceived physical demand and comfort.
We found a significant difference in question 2 (3HANDS median = 2, baseline median = 4, W = 10855, p < .001), indicating 3HANDS allows for generative models for motions with less perceived demands.
Similarly, we found a significant effect for question 7 (3HANDS median = 6, baseline median = 5, W = 3538.5, p < .001) which indicates that 3HANDS led to higher perceived comfort.

\subsubsection{Predictability and Smoothness}

Question 4 (\emph{The motion of the robotic arm was predictable}) and question 5 (\emph{The limb’s movements were smooth and continuous}) examined the perceived predictability and motion smoothness.
Wilcoxon Signed Rank Tests showed no significant differences between 3HANDS and baseline (both $p>0.05$).
Medians of 5 for baseline and 6 for 3HANDS indicate promising handover smoothness and predictability for both approaches.

\subsubsection{Perceived Timeliness and Appropriateness}

Finally, question 6 (\emph{The handover timing matched my expectations}) and question 8 (\emph{The robotic limb reacted quickly and appropriately to my actions}) validated the perceived timeliness and appropriateness of the handover motion.
For question 6, a significant difference was found between 3HANDS and baseline (3HANDS median = 6, baseline median = 5, W = 5512.5, p < .01).
We also found a significant difference in question 8 (3HANDS median = 6, baseline median = 5, W = 5170.5, p < .001).
These findings highlight that the data-driven method is better in timing and more appropriate motion.

\subsubsection{Summary}
To conclude, our approach generated more natural, more comfortable, more timely, and more appropriate handover interaction motions compared to the baseline. The results highlight the application opportunities of our 3HANDS dataset and the presented generative models. 

\section{Discussion and Limitations}

Our results demonstrate that the dataset effectively captures the key features of human-to-human, asymmetric, and asynchronous handover motions, making it well-suited for training SRLs. Notably, our dataset enables models to accurately generate handover motions, predict the handover region, and determine the timing of the handover event.
The positive results in the user study indicated the models trained with the 3HANDS dataset generally result in more natural, more comfortable, and smoother handover interactions. 
Future research should explore more complex architectures.

An important future direction is to implement our model in practical applications. 
Future research should deploy our data-driven handover models in physical SRLs. One potential challenge is bridging the gap between synthetic environments and real-world conditions; exploring various techniques for improving simulation-to-reality (sim2real) transfer will be required. Additionally, rapid and robust control methods coupled with accurate motion sensors may be essential to match the generated physical motions with the desired outcomes. Moreover, the development of safety-aware generative models is critical to ensure that predicted trajectories are compatible with safe and reliable robot operation. 

Our dataset has further potential for applications beyond the models we explored in this paper.
One promising avenue is to analyze human-to-human handover dynamics when primary activities are performed simultaneously with implicit handover requests. 
Research could shed new light on the intricacies of human communication and collaboration by examining how humans coordinate and complete tasks while handing objects to others. 
This understanding could be utilized in various contexts, such as developing more natural and intuitive interfaces for multi-user scenarios. 
Furthermore, our dataset could also serve as a basis for generating realistic human-to-avatar handovers in virtual environments. An exciting and timely direction would be the deployment of our models in virtual reality, where virtual arms could perform object handovers with human users. This application reduces the need for the precise control methods required in the physical world but presents the challenge of rendering realistic motions for objects of varying and unconstrained properties and from different directions. By incorporating the captured nuances in avatar interactions, we can create more immersive and engaging virtual environments that better reflect social dynamics. Increasing the predictability of object interactions from reliable real world data furthermore helps reduce the cognitive load during VR interactions in the absence of additional feedback modalities (like haptics) that are available in real-world SRL interactions. Addressing this challenge will be key to expanding the model's use in virtual environments. 
Additionally, our dataset may be leveraged to train the arm motion of mobile robots to exhibit human-like motion patterns around and near humans, enabling them to navigate complex social situations more easily and naturally.

While our study successfully demonstrates the feasibility of recording human motions for robotic trajectory generation, we acknowledge that the setup exhibits a downward bias in handover locations and human enactment may not always fully capture the nuances of robotic motion, particularly due to the differences in morphology and origin between another person's arm and a specific implementation of a body-worn robotic arm. 
Moreover, the social dynamics of two interacting humans may not be fully representative of interactions with a robotic arm, as the latter might be perceived as inherent extensions of one's own body. We attempted to mitigate this issue to the extent possible by only recording data from couples who live in a stable relationship and hence interact comfortably in close peripersonal space, and by shielding the robot participant's head from the primary user's view.

Future research should consider using our models as pre-trained baselines for fine-tuning in specific applications. For example, researchers could fine-tune our model for handover motions in different robotic arm settings (e.g., shoulder-mounted or environment-mounted). 
%

%
Additionally, our dataset focuses on trajectories without including object-specific details. Different objects may require distinct handling strategies, affecting parameters such as grasp, object orientation, and motion speed. Future work should fine-tune the model with object properties so that it can generate motions tailored to the affordances of different objects, such as mugs containing liquid or heavy items.
Because our models are trained on general handover tasks, they support transfer learning with minimal data. 
While incorporating object affordances and hand interactions are two possible extensions to the handover space defined by 3HANDS, their coverage is out of the main focus of this work. We acknowledge their potential for future work and have included hand joints tracking in the 3HANDS to support future exploration.
Furthermore, our user study on perceived motion quality is deployed in a VR setting to ensure the effects of the confounding variables raised by physical environments are minimized. 
Yet, future research should integrate sophisticated control methods to bring such handover interactions from VR to the physical realm. 

Moreover, our positive outcomes suggest that subsets of our dataset can be used to train lightweight models. For instance, researchers could focus on the most critical joints identified in our experiments (e.g., left hand, left wrist) and develop models accordingly. By gathering additional data on these key joints, transfer learning can be further applied to specific tasks.
Finally, our joint importance analysis offers valuable insights for future data collection in other handover scenarios. In situations where full-body tracking is not feasible, this analysis can guide sensor placement decisions to optimize data collection and deployment of interactive systems.


Finally, a significant and novel challenge we identified is that the autoregressive generation of robot motions does not closely align with ground truth data. 
This challenge arises because the model generates the robot’s motions while observing only the primary user's movements—an issue that significantly differs from existing motion generation tasks. Addressing this challenge may necessitate more complex models.
One potential solution is to enhance the model's reasoning capabilities, allowing it to better interpret the primary user’s intentions (e.g., waiting to receive an object, moving to a destination, being occupied, etc.) and use this additional context to inform the motion generation. Another approach could be the integration of reinforcement learning, which could train a policy model to adapt to the environment and the primary user.

\section{Conclusion}

In this paper, we have presented the \dataset{} dataset, which provides extensive capture of object handovers between closely interacting humans. It considerably extends beyond prior datasets by its asymmetric spatial configuration with handovers occurring in intimate peripersonal space, the participant’s asymmetric roles, real-world primary activities, and implicit coordination of handover. This is representative of the unique demands of handovers between humans and wearable robotic limbs. 12 unique pairings of participants were captured in 41 synchronized 2K camera views, from which we calculate rigged 3D skeleton data and hand poses. The dataset also includes transcripts of utterances, such as verbal commands and reactions, as well as manually annotated ground-truth data for object handover. 

In a series of experiments, we demonstrate the applicability of the dataset for training models for interaction with SRLs. We contribute models and corresponding technical evaluation results that each address one key aspect of a handover activity. We contribute a generative model, based on a conditional variational autoencoder, which generates the trajectory of a handover in response to the primary user’s motion. Furthermore, we present a model that can accurately generate the region of transfer, where an object will be handed over. Additionally, we show that using our dataset it is possible to accurately predict, solely from implicit user posture, when the handover should be initiated. 
Finally, we deployed our models for performing handover interactions compared to an established baseline approach in a VR setting. The user study showed that our data-driven approach enables more natural and comfortable handover interaction, further highlighting the potential value of \dataset{} for training SRL models.

We share the dataset with the community to foster future research on interactive systems and to help deepen the understanding of the unique characteristics of handover activities in close personal space.

\bibliographystyle{ACM-Reference-Format}
\bibliography{main}
\end{document}